\documentclass[twoside,11pt]{article}

\usepackage{subcaption}
\usepackage{jmlr2e}

\usepackage{lastpage}
\jmlrheading{21}{2020}{1-\pageref{LastPage}}{6/19; Revised
2/20}{2/20}{19-468}{Rudy Bunel, Jingyue Lu, Ilker Turkaslan, Philip H.S. Torr, Pushmeet Kohli and M. Pawan Kumar}
\ShortHeadings{Branch and Bound for PL-NN Verification}{Bunel, Lu, Turkaslan, Torr, Kohli and Kumar}

\firstpageno{1}

\usepackage{array}
\usepackage{algorithm}
\usepackage{algpseudocode}
\usepackage{amsmath}
\usepackage{amssymb}
\usepackage{bbold}
\usepackage{bm}
\usepackage{colortbl}
\usepackage{graphicx}
\usepackage{multirow}
\usepackage{wrapfig}
\usepackage{tikz}
\usetikzlibrary{shapes, positioning, arrows,calc}
\newcommand{\mtt}[1]{\mathtt{#1}}
\newcolumntype{?}{!{\vrule width 1pt}}
\usepackage[export]{adjustbox}

\usepackage{placeins}

\DeclareMathOperator*{\for}{for}

\newcommand{\xhat}{\hat{x}}
\newcommand{\mbf}[1]{\mathbf{#1}}

\usepackage[utf8]{inputenc} 
\usepackage[T1]{fontenc}    
\usepackage{url}            
\usepackage{booktabs}       
\usepackage{amsfonts}       
\usepackage{nicefrac}       
\usepackage{microtype}      
\usepackage{natbib}

\makeatother
\begin{document}

\title{Branch and Bound for Piecewise Linear Neural Network Verification}

\author{\name Rudy Bunel \thanks{equal contribution} \email rudy@robots.ox.ac.uk\\
        \name Ilker Turkaslan \email ilker.turkaslan@lmh.ox.ac.uk\\
        \name Philip H.S. Torr \email philip.torr@eng.ox.ac.uk\\
        \name M. Pawan Kumar \email pawan@robots.ox.ac.uk\\
       \addr Department of Engineering Science\\
       University of Oxford\\
       Oxford OX1 3PJ
       \AND
       \name Jingyue Lu \footnotemark[1] \email jingyue.lu@spc.ox.ac.uk \\
       \addr Department of Statistics\\
       University of Oxford\\
       Oxford OX1 3LB
       \AND
       \name Pushmeet Kohli \email pushmeet@google.com\\
       \addr Deepmind\\
       London N1C 4AG
       }

\editor{Amir Globerson}
\maketitle

\begin{abstract}%
  The success of Deep Learning and its potential use in many safety-critical
  applications has motivated research on formal verification of Neural Network
  (NN) models. In this context, verification involves proving or disproving that an NN model satisfies certain input-output properties.
  Despite the reputation of learned NN models as black
  boxes, and the theoretical hardness of proving useful properties about them, researchers
  have been successful in verifying some classes of models by exploiting their
  piecewise linear structure and taking insights from formal methods such as
  Satisifiability Modulo Theory. However, these methods are still far from
  scaling to realistic neural networks. To facilitate progress on this crucial
  area, 
  we exploit the Mixed Integer Linear Programming (MIP) formulation of verification to propose a family of algorithms based on Branch-and-Bound (BaB). We show that our family contains previous verification methods as special cases. With the help of the BaB framework, we make three key contributions. Firstly, we identify new methods that combine the strengths of multiple existing approaches, accomplishing significant performance improvements over previous state of the art. Secondly, we introduce an effective branching strategy on ReLU non-linearities. This branching strategy allows us to efficiently and successfully deal with high input dimensional problems with convolutional network architecture, on which previous methods fail frequently. Finally, we propose comprehensive test data sets and benchmarks which includes a collection of previously released testcases. We use the data sets
  to conduct a thorough experimental comparison of existing and new algorithms and to provide
  an inclusive analysis of the factors impacting the hardness of verification problems.
\end{abstract}


\section{Introduction}
Despite their success in a wide variety of applications, Deep neural networks
have seen limited adoption in safety-critical settings. The main explanation for
this lies in their reputation for being black-boxes whose behaviour cannot be
predicted. Current approaches to evaluate trained models mostly rely on testing
using held-out data sets. However, as Edsger W. Dijkstra
said ``testing shows the presence, not the absence of
bugs''~\citep{buxton1970}. If deep learning models are to be deployed in applications such as
autonomous driving cars, we need to be able to verify safety-critical
behaviours.
\par
To this end, some researchers have tried to use formal methods. To the best of
our knowledge, \citet{Zakrzewski2001} was the first to propose a method to
verify simple, one hidden layer neural networks. However, only recently were
researchers able to work with non-trivial models by taking advantage of the
structure of ReLU-based networks~\citep{Cheng2017a,Katz2017}. Even then, these
works are not scalable to the large networks encountered in most real world
problems.
\par
This paper advances the field of NN verification by making the following key
contributions:
\begin{enumerate}
\item By taking advantage of the Mixed Integer Linear Programming (MIP) formulation of the problem, we introduce the Branch-and-Bound framework for NN verification. The framework contains state of the art verification methods as special cases.
\item We identify the weakness and strengths of previous verification methods from the aspects of the way bounds are computed, the type of
  branching that are considered and the strategies guiding the
  branching. By retaining the strengths and correcting the identified flaws, we propose new methods that achieve considerable performance improvements when compared to the previous state of the art. In some cases, a speed-up of almost two orders of magnitudes is obtained. Specifically, we develop a new branching strategy that supports branching over ReLU non-linearities. Previous BaB based verification methods mainly focus on designing heuristics for branching over input domains. These heuristics, although they perform well on small-scale problems, are either computationally expensive for high dimensional input problems or ineffective for problems with convolutional network architecture. Similar issues are faced by the existing ReLU branching strategies. Our designed branching strategy is computationally cheap and explores the underlying network architecture to make a decision. Using on high dimensional input problems with convolutional network architectures, we demonstrate the benefits of our branching strategy over various verification methods that employ either input-domain branching or ReLU branching strategies. 
\item We introduce comprehensive data sets consisting of trained as well as synthetic networks with fully connected and/or convolutional layers. Convolutional networks are widely used in computer vision tasks and should be an indispensable component for fair and complete evaluations of verification methods. Only recently did convolutional network data start to be included in the evaluation of verification methods. This takes the form of verification properties attempting to prove adversarial robustness on a $\mathcal{L}_{\infty}$ ball with a fixed perturbed distance $\epsilon$. The difficulty level of a verification property is mainly determined by its network size. Our curated convolutional data sets differ from these data sets and are able to bring new insights by verifying properties on range of $\epsilon$ values on the same network. We make two observations to strengthen our statement. Firstly, the difficulty of a verification property not only relies on the size of the network, but also the value of $\epsilon$. Secondly, one bottleneck for BaB based methods is the time required for solving linear programs (LPs), which could increase significantly with the size of the network. Our data sets consist of verification properties with various difficulty levels on relatively small network architecture. This means they allow effective evaluations of branching heuristics or bounding decisions without suffering from the LP bottleneck. Additionally, we have introduced the synthetic TwinStream data set to facilitate the study of the relationship between bounding and branching strategies. Overall, the extensive test data sets not only allow thorough experimental analyses of existing methods, but also facilitate the understanding of verification problems and encourage the development of new methods. 
\end{enumerate}
A preliminary version of this work appeared in the proceedings of NeurIPS, 2018~\citep{Bunel2018}. The article significantly differs from the previous work by (i) improving the clarity of the BaB framework by providing a running toy example; (ii) designing novel branching strategies for the important class of NN with convolutional layers; (ii) introducing new data sets with convolutional networks and synthetic models; and (iv) including new baseline algorithms.
\par

Section 2 and 3 specify the problem of verification and present different formulations of
verification processes respectively. Section 4 presents the BaB framework, showing that previous methods can be seen as special cases of it. Section 5 builds on the observations in section 4 to highlight possible improvements within the BaB framework. New methods are proposed accordingly. The last two sections conduct detailed experimental studies of verification methods on our comprehensive data sets. Specifically, section 6 discusses the experimental setup and section 7 analyses the results.

\section{Problem Specification}
We now specify the problem of formal verification of neural networks. Given a
network that implements a function $\mbf{\xhat_n} = f(\mbf{x_0})$, a bounded input domain
$\mathcal{C}$ and a property $P$, we want to prove
\begin{equation}
  \mbf{x_0} \in \mathcal{C},\quad \mbf{\xhat_n} = f(\mbf{x_0}) \implies P(\mbf{\xhat_n}).
  \label{eq:prob-form}
\end{equation}
For example, the property of robustness to adversarial examples in
$\mathcal{L}_{\infty}$ norm around a training sample $\mbf{a}$ with label $y_a$ would be encoded by
using \mbox{$\mathcal{C} \triangleq \left\{\mbf{x_0} | \ \|\mbf{x_0} -
      \mbf{a}\|_{\infty} \leq \epsilon\right\}$} and
  \mbox{$P(\mbf{\xhat_n}) =\left\{ \forall y, \quad \xhat_{n
        [y_a]} > \xhat_{n [y]}\right\}$}. From now on, we use $x_{i[j]}$ to denote the $j$th element of $\mbf{x_i}$.
\par
  In this paper, we are going to focus on Piecewise-Linear neural networks
  (PL-NN), that is, networks for which we can decompose $\mathcal{C}$ into a set
  of polyhedra $\mathcal{C}_i$ such that \mbox{$\mathcal{C} = \cup_i\
    \mathcal{C}_i$}, and the restriction of $f$ to $\mathcal{C}_i$ is a linear
  function for each $i$. While this prevents us from including networks that use
  activation functions such as sigmoid or tanh, PL-NNs allow the use of linear
  transformations such as fully-connected or convolutional layers, pooling units
  such as MaxPooling and activation functions such as ReLUs. In other words,
  PL-NNs represent the majority of networks used in practice. Operations such as
  Batch-Normalization or Dropout also preserve piecewise linearity at
  test-time. 
\par
The properties that we are going to consider are Boolean formulas over linear
inequalities. In our robustness to adversarial example above, the property is a
conjunction of linear inequalities, each of which constrains the output of the
original label to be greater than the output of another label.
\par
In general, we divide verification algorithms into three categories: algorithms are unsound if they can only prove some of the false properties are false; algorithms are incomplete if they can only prove some of the true properties are true; and algorithms are complete if they are able to report all correct properties. In this paper, we will only focus on complete algorithms. For unsound methods, we refer interested readers to \citet{NSVerify,Huang2017}, \citet{Carlini2017} and \citet{Stefan2019} and for incomplete methods, we refer interested readers to \citet{Xiang2017, Weng2018, deepz2018} and \citet{Dvijotham2018}. In addition, among complete methods, the scope of this paper does not include approaches relying on additional
assumptions such as twice differentiability of the network~\citep[see][]{Hein2017, Zakrzewski2001}, limitation of the activation to binary values~\citep[see][]{Cheng2017a,Narodytska2017} or restriction to a single linear
domain~\citep[see][]{Bastani2016}. 

\section{Verification Formalism}
In this section, we present different formulations of verification process.
\subsection{Verification as a Satisfiability Problem}
\label{subsec:verif-as-opt}
The methods we involve in our comparison all leverage the piecewise-linear
structure of PL-NN to make the problem more tractable. They all follow the same
general principle: given a property to prove, they attempt to discover a
counterexample that would make the property false. This is accomplished by
defining a set of variables corresponding to the inputs, hidden units and output
of the network, and the set of constraints that a counterexample would
satisfy.
\par
To help design a unified framework, we reduce all instances of verification
problems to a canonical representation. Specifically, the whole satisfiability problem will be transformed into a global optimization problem where the
decision will be obtained by checking the sign of the minimum. If the property is a simple inequality \mbox{$P(\mbf{\xhat_n}) \triangleq
  \mbf{c}^T \mbf{\xhat_n} > b$}, it is sufficient to add to the network a
final fully connected layer with one output, with weight of $\mbf{c}$ and a bias
of $-b$. If the global minimum of this network is positive, it indicates that for all $\mbf{\xhat_n}$, the original network output, we have
\mbox{$\mbf{c}^T \mbf{\xhat_n} -b > 0 \implies \mbf{c}^T \mbf{\xhat_n} >
  b$}, and as a consequence the property is true. On the other hand, if the
global minimum is negative, then the minimizer provides a counter-example. Clauses \texttt{OR} and \texttt{AND} in the
property can similarly be expressed as additional layers, using MaxPooling units. Specifically, 
clauses specified using \texttt{OR} (denoted by $\bigvee$) can be encoded by
using a MaxPooling unit. If the property is \mbox{$P(\mbf{\xhat_n}) \triangleq
  \bigvee_i\left[ \mbf{c}_i^T \mbf{\xhat_n} > b_i \right]$}, this is
equivalent to $\max_i\left( \mbf{c}_i^T \mbf{\xhat_n} - b_i \right) > 0$. Clauses specified using \texttt{AND} (denoted by $\bigwedge$) can be encoded
similarly: the property $P(\mbf{\xhat_n}) = \bigwedge_i \left[
  \mbf{c}_i^T\mbf{\xhat_n} > b_i \right]$ is equivalent to
\mbox{\small$\min_i \left( \mbf{c}_i^T \mbf{\xhat_n} - b_i \right) > 0 \iff
  -\left( \max_i\left(-\mbf{c}_i^T \mbf{\xhat_n} + b_i \right) \right) > 0$}.
We can formulate any Boolean formula over linear inequalities on the output of
the network as a sequence of additional linear and max-pooling layers. From now on, we 
assume that a property is in a canonical form. Specifically, the output of the network is 
a scalar, and the property is true if the output is positive for all inputs in a given domain, and
false otherwise. Assuming the network only contains ReLU activations between each layer, the
satisfiability problem to find a counterexample can be expressed as:
\begin{subequations}
  \begin{minipage}{.25\textwidth}
    \begin{align}
    &\mbf{l_0} \leq \mbf{x_0} \leq \mbf{u_0} \label{eq:ctx-bounds0}\\
    &\xhat_{n} \leq 0 \label{eq:final0}
    \end{align}
  \end{minipage}%
  \begin{minipage}{.75\textwidth}
    \begin{align}
      \qquad&\mbf{\xhat_{i+1}} = W_{i+1} \mbf{x_i} + \mbf{b_{i+1}} \qquad&&\forall i \in \{0, \ n-1\}\label{eq:lin-op0}\\
      \qquad&\mbf{x_i} = \max\left(\mbf{\xhat_i}, 0\right) \ \ \quad\qquad&&\forall i \in \{1, \ n-1\} .\label{eq:relu-op0}
    \end{align}
  \end{minipage}
  \label{eq:sat-problem}
\end{subequations}

Equation~\ref{eq:ctx-bounds0} represents the constraints on the input and
Equation~\ref{eq:final0} on the neural network output.
Equation~\ref{eq:lin-op0} encodes the linear layers of the network and
Equation~\ref{eq:relu-op0} the ReLU activation functions. If an assignment to all
the values can be found, this represents a counterexample. If this problem is
unsatisfiable, no counterexample can exist, implying that the property is true.
We emphasise that we are required to prove that no counter-examples can exist,
and not simply that none could be found.
\par
While for clarity of explanation, we have limited ourselves to the specific case
where only ReLU activation functions are used, this is not restrictive. The
appendix contains a section detailing how each method specifically
handles MaxPooling units, as well as how to convert any MaxPooling operation into
a combination of linear layers and ReLU activation functions. Converting a verification problem into this canonical
representation does not make its resolution simpler since the addition of the ReLU non-linearities Equation~\ref{eq:relu-op0} transforms a problem
that would have been solvable by simple Linear Programming into an NP-hard
problem~\citep{Katz2017}. However, it does provide a
formalism advantage. Specifically, it allows us to prove complex properties,
containing several \texttt{OR} clauses, with a single procedure rather than
having to decompose the desired property into separate queries as was done in
previous work~\citep{Katz2017}. Operationally, a valid strategy for dealing with verification problems in the canonical form is to impose the constraints
Equations~\ref{eq:ctx-bounds0}-\ref{eq:relu-op0} and minimise the value of
$\xhat_{n}$. Finding the exact global minimum is not necessary for verification.
However, it provides a measure of satisfiability or unsatisfiability. If the
value of the global minimum is positive, it will correspond to the margin by
which the property is satisfied.
\par
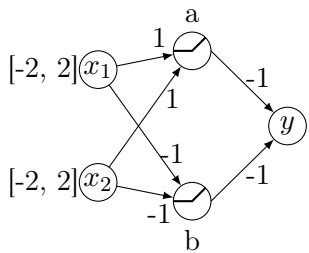
\begin{figure}
  \centering
  \begin{tikzpicture}

  \tikzset{dummy/.style= {inner sep=0, outer sep=0}}
  \tikzset{neuron/.style={draw, circle,inner sep=0, outer sep=0, minimum size=0.5cm}};

  \node[neuron](x) {$x_1$};
  \node[dummy, left = 0 of x](x_range) {[-2, 2]};
  \node[neuron, below = of x](y) {$x_2$};
  \node[dummy, left = 0 of y](y_range) {[-2, 2]};

  \node[neuron, right = of x.north](a) {};
  \draw[black, thick](a.west) to (a.center) to (a.north east);
  \node[above = 0 of a](alabel) {a};
  \node[neuron, right = of y.south](b) {};
  \draw[black, thick](b.west) to (b.center) to (b.north east);
  \node[below = 0 of b](blabel) {b};

  \draw[-latex] (x) to node[near end, above](Wxa){1} (a);
  \draw[-latex] (y) to node[very near end, below](Wya){1} (a);
  \draw[-latex] (x) to node[very near end, above](Wxb){-1} (b);
  \draw[-latex] (y) to node[near end, below](Wyb){-1} (b);

  \node[dummy](inp-mid) at ($(a)!0.5!(b)$){};
  \node[neuron, right = of inp-mid](out) {$y$};

  \draw[-latex] (a.east) to node[near end, above](Wao){-1} (out);
  \draw[-latex] (b.east) to node[near end, below](Wbo){-1} (out);

  \node[draw, thick, below = of b](property) {\textbf{Prove that $y > -5$}};
\end{tikzpicture}
  \caption{\label{fig:net-example}  Example Neural Network. We attempt to prove
    the property that the network output is always greater than -5.}
\end{figure}

\textbf{Toy Example}\textit{
A toy-example of the Neural Network verification problem is given in
Figure~\ref{fig:net-example}. On the domain $\mathcal{C} = [-2; 2] \times [-2;
2]$, we want to prove that the output $y$ of the one hidden-layer network
always satisfies the property $P(y) \triangleq \left[ y > -5 \right]$. We will use
this as a running example to illustrate different formulations of the problem and introduce methods that can be reframed in our unified framework.}

\textit{For the network of Figure~\ref{fig:net-example}, the problem is formulated as follows. The variables would be $\{x_1,
x_2, a_\text{in}, a_\text{out}, b_\text{in}, b_\text{out}, y\}$ and the set of
constraints would be:}

\begin{subequations}
  \begin{align}
    & -2 \leq x_1 \leq 2 \qquad&& -2 \leq x_2 \leq 2  \label{eq:ex-bounds}\\
    & \hat{a} = x_1 + x_2  \qquad&& \hat{b} = -x_1 - x_2  \label{eq:ex-lin}\\
    & a = \max(\hat{a}, 0) \qquad&& b = \max(\hat{b}, 0)\label{eq:ex-nonlin}  \\
    &   y = -a - b\\
    &y \leq -5.   \label{eq:ex-final}
  \end{align}
\end{subequations}

\textit{Here, $\hat{a}, \hat{b}$ is the input value to hidden unit, while $a, b$ is the value after the ReLU. Any point satisfying all the above constraints would be a
counterexample to the property, as it would imply that it is possible to drive
the output to -5 or less.}


\subsection{Mixed Integer Linear Programming Formulation}
\label{subsec:mip}
A possible way to eliminate the non-linearities is to encode them with the help
of binary variables, transforming the PL-NN verification
problem Equation~\ref{eq:sat-problem} into a Mixed Integer Linear Programming (MIP) problem. This
can be done with the use of ``big-M'' encoding. The following encoding is from
\citet{Tjeng2019}. Assuming we have access to lower and upper bounds on the
values that can be taken by the coordinates of $\mbf{\xhat_i}$, which we denote
$\mbf{l_i}$ and $\mbf{u_i}$, we can replace the non-linearities:
\begin{subequations}
  \begin{flalign}
    \mbf{x_i} = \max\left(\mbf{\xhat_i}, 0\right) \quad \Rightarrow \quad \bm{\delta_i} \in
    \{0,1\}^{h_i}, &\quad\mbf{x_i} \geq 0, \ \qquad  \mbf{x_i} \leq \mbf{u_i} \cdot \bm{\delta_i}\label{eq:mip_0}\\
    & \quad \mbf{x_i}\geq \mbf{\xhat_i},   \qquad \mbf{x_i} \leq \mbf{\xhat_i} - \mbf{l_i}\cdot(1 - \bm{\delta_i}).\label{eq:mip_inp}
  \end{flalign}
  \label{eq:mip-form0}%
\end{subequations}
It is easy to verify that \mbox{$\delta_{i[j]}=0 \Leftrightarrow x_{i[j]}=0$}
(replacing $\delta_{i[j]}$ in Equation~\ref{eq:mip_0}) and $\delta_{i[j]}=1
  \Leftrightarrow x_{i[j]}=\xhat_{i[j]}$ (replacing $\delta_{i[j]}$ in
Equation~\ref{eq:mip_inp}). From now on, we refer to $\mbf{l_0}$ and $\mbf{u_0}$, the lower and upper bounds for the input domain, as input bounds and $\mbf{l_i}$ and $\mbf{u_i}$ for $i\in  \{1, \ n-1\}$, the lower and upper bounds for hidden units $\hat{\mbf{x_i}}$, as intermediate bounds.
\par
\textbf{Toy Example (MIP formulation)}
\textit{In our example, the non-linearities of Equation~\ref{eq:ex-nonlin} would be
replaced by following conditions. Here, we give the detailed description for $\hat{a}$. The same is applied to $\hat{b}$.}
  \begin{align}
    &a \geq 0 \qquad\qquad&&a \geq \hat{a} \nonumber \\
    &a \leq \hat{a} - l_a (1-\delta_a) \qquad\qquad&& a \leq u_a \delta_a \\
    &\delta_a \in \{0, 1\}. \nonumber
  \label{eq:mip-relu}
  \end{align}
\textit{where $l_a$ is a lower bound of the value that $\hat{a}$ can take (such as -4)
and $u_a$ is an upper bound (such as 4). The binary variable $\delta_a$
indicates which phase the ReLU is in: if $\delta_a=0$, the ReLU is blocked and
$a=0$, else the ReLU is passing and $a = \hat{a}$. The
problem remains difficult due to the integer constraint on $\delta_a$.}
\par
By taking advantage of the feed-forward structure of the neural network, lower
and upper bounds $\mbf{l_i}$ and $\mbf{u_i}$ can be obtained by applying
interval arithmetic~\citep{Hickey2001} to propagate the bounds on the inputs, one
layer at a time.
\par
Thanks to this specific feed-forward structure of the problem, the generic,
non-linear, non-convex problem has been rewritten into a MIP. Optimization of
MIP is well studied and highly efficient off-the-shelf solvers exist. As solving
them is NP-hard, performance is going to be dependent on the quality of both the
solver used and the encoding. We now ask the following question: how much
efficiency can be gained by using a bespoke solver rather than a generic one? In
order to answer this, we present specialised solvers for the PL-NN verification
task.


\section{Branch and Bound for Verification}
As described in Section~\ref{subsec:verif-as-opt}, the verification problem can
be rephrased as a global optimization problem. For non-convex problems, algorithms such as
first order methods like Gradient Descent are not appropriate as they have no way of
guaranteeing whether or not a stationary point is a global minimum. In this section, we present an
approach to estimate the global minimum, based on the Branch-and-Bound paradigm~\citep{bab1960, babsurvey}. We also show that several published methods fit this framework. Among these methods, detailed studies are conducted on the previous state of art methods Reluplex and Planet, which are introduced as examples of Satisfiability Modulo Theories (SMT). 
\par
\begin{wrapfigure}[22]{L}{.56\textwidth}
  \begin{minipage}{.56\textwidth}
    \vskip -20pt
    \begin{algorithm}[H]
      \caption{Branch and Bound}\label{alg:bab}
      \small
      \begin{algorithmic}[1]
        \algrenewcommand\algorithmicindent{1.0em}%
        \Function{BaB}{$\mtt{net}, \mtt{problem}, \epsilon$}
        \State $\mtt{global\_ub} \gets \inf$
        \State $\mtt{global\_lb} \gets - \inf$
        \State $\mtt{probs}\gets \left[ (\mtt{global\_lb}, \mtt{problem}) \right]$
        \While{$\mtt{global\_ub} - \mtt{global\_lb} > \epsilon $}
        \State $(\_\ , \mtt{prob}) \gets \mtt{pick\_out}(\mtt{probs})$
        \State $\left[ \mtt{subprob\_1}, \dots, \mtt{subprob\_s} \right] \gets \mtt{split}(\mtt{prob})$
        \For{$i = 1 \dots s $}
        \State $\mtt{prob\_ub} \gets \mtt{compute\_UB}(\mtt{net}, \mtt{subprob\_i})$
        \State $\mtt{prob\_lb} \gets \mtt{compute\_LB}(\mtt{net}, \mtt{subprob\_i})$
        \If{$\mtt{prob\_ub} < \mtt{global\_ub}$}
        \State $\mtt{global\_ub} \gets \mtt{prob\_ub}$
        \State $\mtt{prune\_problems}(\mtt{probs}, \mtt{global\_ub})$
        \EndIf
        \If{$\mtt{prob\_lb} < \mtt{global\_ub}$}
        \State $\mtt{problems}.\mtt{append}((\mtt{prob\_lb}, \mtt{subprob\_i}))$
        \EndIf
        \EndFor
        \State $\mtt{global\_lb} \gets \min\{ \mtt{lb}\ |\ (\mtt{lb}, \mtt{prob}) \in \mtt{probs}\}$
        \EndWhile
        \State\Return $\mtt{global\_ub}$
        \EndFunction
      \end{algorithmic}
    \end{algorithm} 
  \end{minipage}

\end{wrapfigure}
Algorithm~\ref{alg:bab} describes its generic form. The original problem is
repeatedly split into sub-problems (either split the input domain into sub-domains or an unfixed ReLU activation unit \footnote{We refer to a ReLU activation unit $x_i[j] = \max(\hat{x}_{i[j]}, 0)$ as unfixed if, given the upper and lower bounds $u_{i[j]}, l_{i[j]}$ of $x_{i[j]}$, $x_i[j]$ can take either the value of $\hat{x}_{i[j]}$ or $0$. } into different phases) (line 7), over which lower and upper bounds of
the minimum are computed (lines 9-10). The best upper-bound found so far serves
as a candidate for the global minimum. Any sub-problem whose lower bound is greater
than the current global upper bound can be pruned away as it cannot contain the
global minimum (line 13, lines 15-17). By iteratively splitting the (sub-)problems, it
is possible to compute tighter lower bounds. We keep track of the global lower
bound on the minimum by taking the minimum over the lower bounds of all
sub-problems (line 19). When the global upper bound and the global lower bound
differ by less than a small scalar $\epsilon$ (line 5), we consider that we have
converged.
\par
Algorithm~\ref{alg:bab} shows how to optimise and obtain the global minimum.
If all that we are interested in is the satisfiability problem, the procedure
can be simplified by initialising the global upper bound with 0 (in line 2). Any
sub-problem with a lower bound greater than 0 (and therefore not eligible to
contain a counterexample) will be pruned out (by line 15). The computation of
the lower bound can therefore be replaced by the feasibility problem (or its
relaxation) imposing the constraint that the output is below zero without
changing the algorithm. If it is feasible, there might still be a counterexample
and further branching is necessary. If it is infeasible, the sub-problem can be
pruned out. In addition, if any upper bound improving on 0 is found on a
sub-problem (line 11), it is possible to stop the algorithm as this already
indicates the presence of a counterexample.

The description of the verification problem as optimization and the pseudo-code
of Algorithm~\ref{alg:bab} are generic and would apply to verification problems
beyond the specific case of PL-NN. To obtain a practical algorithm, it is
necessary to specify several elements.

{\bf A search strategy}, defined by the $\mtt{pick\_out}$ function, which
chooses the next problem to branch on. Several heuristics are possible, for
example those based on the results of previous bound computations. For
satisfiable problems or optimization problems, this allows us to discover good
upper bounds, enabling early pruning.

{\bf A branching rule}, defined by the $\mtt{split}$ function, which takes a problem $\mtt{prob}$ and returns its partition into subproblems such that
\mbox{$\bigcup_i \mtt{subprob\_i} = \mtt{prob}$} and that\\
\mbox{$\left(\mtt{subprob\_i} \cap \mtt{subprob\_j} \right)= \emptyset, \ \forall i \neq j$}. This will determine the attributes of the (sub-)problems, which impacts the hardness
of computing bounds. In addition, choosing the right partition can greatly
impact the quality of the resulting bounds.

{\bf Bounding methods}, defined by the $\mtt{compute\_\{UB,LB\}}$
functions. These procedures estimate respectively upper bounds ($\mtt{prob\_ub}$) and lower bounds ($\mtt{prob\_ub}$)
over the minimum output that the network $\mtt{net}$ can reach over a given (sub-)problem. We want the lower bound to be as high as possible, so that this
(sub-)problem can be pruned easily. This is usually done by introducing convex
relaxations of the (sub-)problem and minimising them. On the other hand, the computed upper bound should be as small as possible, so as to allow pruning out other (sub-)problems or discovering counterexamples. As any feasible point
corresponds to an upper bound on the minimum, heuristic methods are sufficient.

\subsection{BaB Reformulations}
We now give a general discussion of published work in the verification literature through the lens of the Branch-and-Bound framework. We first briefly mention some methods that do not rely on SMT solvers and then conduct detailed studies on SMT based methods Reluplex and Planet. ReluVal is a complete method introduced by \citet{wang2018formal}. In ReluVal, an input domain branching rule is used and the split function decides which domain dimension to split on by computing influence metrics based on input-output gradient. For the bounding method, it uses a novel technique called symbolic interval propagation, which replaces lower and upper bounds by linear equations of input variables and propagates these linear equations in a layer by layer order. By doing so, symbolic intervals can preserve more dependency information than interval arithmetic, and are thus able to achieve tighter final bounds. Inspired by the branching rule used by ReluVal, \citet{royo2019fast} proposed a verification procedure via shadow prices. This method also adopts a Branch-and-Bound framework. In detail, for the branching rule, the method makes an input split decision through sensitivity studies of the bounds of unfixed ReLU nodes to the change of the input domain. The bounding method is not specified but the same logic has been used to check whether a sub-domain should be stored and further split on or should be pruned. Also based on ReluVal is a complete method called Neurify \citep{Wang2018}. Neurify is an improved version of ReluVal, which also uses gradient metrics to make a split decision but it splits on unfixed ReLU nodes. Neurify uses a different bounding method than ReluVal by calling an LP solver instead of computing symbolic interval bounds. For all the methods mentioned above, no specific search strategy is implemented. All sub-problems are simply enumerated. 

\subsection{Reluplex}
\label{subsec:reluplex}
\citet{Katz2017} present a procedure named Reluplex to verify properties of
Neural Network containing linear functions and ReLU activation units. Reluplex functions as an SMT solver using the splitting-on-demand framework~\citep{Barrett2006}. The
principle of Reluplex is to always maintain an assignment to all of the
variables, even if some of the constraints are violated.
\par
Starting from an initial assignment, it attempts to fix some violated
constraints at each step. It prioritises fixing linear constraints
(Equations~\ref{eq:ctx-bounds0}, \ref{eq:final0}, \ref{eq:lin-op0} and some relaxation
of Equation~\ref{eq:relu-op0}) using a simplex algorithm, even if it leads to violated
ReLU constraints. If no solution to this relaxed problem containing only linear
constraints exists, the counterexample search is unsatisfiable. Otherwise,
either all ReLU are respected, which generates a counterexample, or Reluplex
attempts to fix one of the violated ReLU, potentially leading to newly violated
linear constraints. This process is not guaranteed to converge. Thus, to make
progress, non-linearities that get fixed too often are split into two cases. Two
new problems are generated, each corresponding to one of the phases of the ReLU.
In the worst setting, the problem will be split completely over all possible
combinations of activation patterns, at which point the sub-problems will all be
simple LPs.
\par
This algorithm is a special case of Branch-and-Bound for
satisfiability. The {\bf search strategy} is handled by the SMT core and to the
best of our knowledge does not prioritise any (sub-)problems. The {\bf branching rule} is
implemented by the ReLU-splitting procedure: when neither the upper bound
search, nor the detection of infeasibility are successful, one non-linear
constraint over the $j$-th neuron of the $i$-th layer \mbox{$x_{i[j]} =
  \max\left( \xhat_{i[j]}, 0 \right)$} is split out into two sub-problems:
\mbox{$\{x_{i[j]}=0, \xhat_{i[j]}\leq 0\}$} and \mbox{$\{x_{i[j]}=\xhat_{i[j]},
  \xhat_{i[j]}\geq 0\}$}. The
prioritisation of ReLUs that have been frequently fixed is a heuristic to decide
among possible partitions.
\par
As Reluplex only deals with satisfiability, the analogue of the lower bound
computation is an over-approximation of the satisfiability problem. The {\bf
  bounding method} used is a convex relaxation, obtained by dropping some of the
constraints. The following relaxation is applied to ReLU units for which the sign of the
input is unknown \mbox{($l_{i[j]} \leq 0$ and $u_{i[j]} \geq 0$)}.
\begin{subequations}
  \begin{minipage}[b]{.47\textwidth}
  \centering
    \begin{equation}
    \mbf{x_i} = \max\left(\mbf{\xhat_i}, 0\right)\quad \Rightarrow \quad  \mbf{x_i} \geq
    \mbf{\xhat_i} \label{eq:rpx-reluout-lbinp} 
    \end{equation}
  \end{minipage}%
  \begin{minipage}[b]{.26\textwidth}
  \centering
    \noindent\begin{equation}
    \mbf{x_i} \geq 0 \label{eq:rpx-reluout-lb0} 
    \end{equation}
  \end{minipage}%
  \begin{minipage}[b]{.26\textwidth}
  \centering
    \noindent\begin{equation}
    \mbf{x_i} \leq \mbf{u_i}. \label{eq:rpx-reluout-ub}
    \end{equation}
  \end{minipage}
  \label{eq:reluplex-relax}%
\end{subequations}
\par
If this relaxation is unsatisfiable, this indicates that the
subdomain cannot contain any counterexample and can be pruned out. The search for
an assignment satisfying all the ReLU constraints by iteratively attempting to
correct the violated ReLUs is a heuristic that is equivalent to the search for an upper
bound lower than 0: success implies the end of the procedure.
\par
\textbf{Toy Example (Running Reluplex)}
\textit {Figure~\ref{tab:reluplex-ex} shows the initial steps of a run of the Reluplex
algorithm on the example of Figure~\ref{fig:net-example}. Starting from an
initial assignment, it attempts to fix some violated constraints at each step.
It prioritises fixing linear constraints (Equations~\ref{eq:ex-bounds},
\ref{eq:ex-lin} and \ref{eq:ex-final} in our illustrative example) using a simplex algorithm, even if it leads to violated ReLU constraints Equation~\ref{eq:ex-nonlin}. This can be seen
in step 1 and 3 of the process.}

\begin{figure}
  \centering
  {\small
    \begin{tabular}{?c?c|c|c|c|c|c|c?}
      \specialrule{1.5pt}{1pt}{1pt}
      \textbf{Step}& $\mbf{x_1}$ & $\mbf{x_2}$ & $\mbf{\hat{a}}$ & $\mbf{a}$ & $\mbf{\hat{b}}$ & $\mbf{b}$ & $\mbf{y}$ \\
      \specialrule{1.5pt}{1pt}{1pt}
      \multirow{3}{1em}{1}&0 & 0 & 0 & 0 & 0 & 0 & \cellcolor{red}0 \\[0.1ex]
      & \multicolumn{7}{c?}{Fix linear constraints}\\
      & 0 & 0 & \cellcolor{orange}0 & \cellcolor{orange}1 & \cellcolor{orange}0 & \cellcolor{orange}4& -5 \\
      \specialrule{1.5pt}{1pt}{1pt}
      \multirow{3}{1em}{2}& 0 & 0 & \cellcolor{orange}0 & \cellcolor{orange}1 & \cellcolor{orange}0 & \cellcolor{orange}4 & -5 \\
      & \multicolumn{7}{c?}{Fix a ReLU}\\
      & \cellcolor{red}0 & \cellcolor{red}0 & \cellcolor{orange}0 & \cellcolor{orange}1 & \cellcolor{red}4 & 4 & -5 \\
      \specialrule{1.5pt}{1pt}{1pt}
      \multirow{3}{1em}{3} & \cellcolor{red}0 & \cellcolor{red}0 & \cellcolor{orange}0 & \cellcolor{orange}1 & \cellcolor{red}4 & 4 & -5 \\
      & \multicolumn{7}{c|}{Fix linear constraints}\\
      & -2 & -2 & \cellcolor{orange}-4 & \cellcolor{orange}1 & 4 & 4 & -5 \\
      \specialrule{1.5pt}{1pt}{1pt}
      & \multicolumn{7}{|c?}{$\dots$}\\
   \end{tabular}
 }
 \caption{ \label{tab:reluplex-ex} Evolution of the Reluplex algorithm. Red
   cells corresponds to value violating linear constraints, and orange cells
   corresponds to value violating ReLU constraints. Resolution of violation of linear
   constraints are prioritised.}
\end{figure}
\par
\textit{If no solution to the problem containing only linear constraints exists, this
shows that the counterexample search is unsatisfiable. Otherwise, all linear
constraints are fixed and Reluplex attempts to fix one violated ReLU at a time,
such as in step 2 of Figure~\ref{tab:reluplex-ex} (fixing the ReLU $b$),
potentially leading to newly violated linear constraints. In the case where no
violated ReLU exists, this means that a satisfiable assignment has been found
and that the search can be terminated early.}

\subsection{Planet}
\label{subsec:planet}
\citet{Ehlers2017} also proposed an approach based on SMT. Unlike Reluplex, the
proposed tool, named Planet, operates by explicitly attempting to find an
assignment to the phase of the non-linearities. Reusing the notation of
Section~\ref{subsec:mip}, it assigns a value of 0 or 1 to each $\delta_{i[j]}$
variable, verifying at each step the feasibility of the partial assignment so as
to prune infeasible partial assignment early.
\par
As in Reluplex, the {\bf search strategy} is not explicitly encoded and simply
iterates over the (sub-)problems that have not yet been pruned. The {\bf branching rule} is the
same as for Reluplex, as fixing the decision variable $\delta_{i[j]}=0$ is
equivalent to choosing \mbox{$\{x_{i[j]}=0, \xhat_{i[j]}\leq 0\}$} and fixing
$\delta_{i[j]}=1$ is equivalent to \mbox{$\{x_{i[j]}=\xhat_{i[j]},
  \xhat_{i[j]}\geq 0\}$}. Note however that Planet does not include any heuristic to prioritise which decision variables should be split over. As a result, there is no mechanism that based on a heuristic search of a feasible point to encourage early termination in Planet.
  
\par
For satisfiable problems, only when a full complete
assignment is identified is a solution returned. In order to detect incoherent
assignments, \citet{Ehlers2017} introduces a global linear approximation to a neural
network, which is used as a {\bf bounding method} to over-approximate the set of
values that each hidden unit can take. In addition to the existing
linear constraints (Equations~\ref{eq:ctx-bounds0}, \ref{eq:final0}and
\ref{eq:lin-op0}), the non-linear constraints are approximated by sets of linear
constraints representing the convex hull of each non-linearity treated independently. Specifically, ReLUs with
input of unknown sign are replaced by the set of equations:
\begin{subequations}
  \begin{minipage}{.42\textwidth}
    \noindent\begin{equation}
      \mbf{x_i} = \max\left(\mbf{\xhat_i}, 0\right)\ \Rightarrow \ \mbf{x_i} \geq \mbf{\xhat_i}
      \label{eq:planet-reluout-lbinp}
    \end{equation}
  \end{minipage}
  \begin{minipage}{.21\textwidth}
    \noindent\begin{equation}
    \mbf{x_i} \geq 0 \label{eq:planet-reluout-lb0}
    \end{equation}
  \end{minipage}
  \begin{minipage}{.36\textwidth}
    \noindent\begin{equation}
      x_{i[j]} \leq u_{i[j]} \frac{\xhat_{i[j]} - l_{i[j]}}{u_{i[j]} - l_{i[j]}}.\label{eq:planet-reluout-ub}
    \end{equation}
  \end{minipage}
  \label{eq:planet-relu-hull}%
\end{subequations}
Recall that $x_{i[j]}$ corresponds to the value of the $j$-th coordinate of
$\mbf{x_{i}}$. An illustration of the convex hull is provided in the
supplementary material.
\par

Compared with the relaxation of Reluplex Equation~\ref{eq:reluplex-relax}, the Planet
relaxation is tighter. Specifically, Equations~\ref{eq:rpx-reluout-lbinp} and
\ref{eq:rpx-reluout-lb0} are identical to Equations~\ref{eq:planet-reluout-lbinp}
and \ref{eq:planet-reluout-lb0} but Equation~\ref{eq:planet-reluout-ub} implies
Equation~\ref{eq:rpx-reluout-ub}. Indeed, given that $\xhat_{i[j]}$ is smaller than
$u_{i[j]}$, the fraction multiplying $u_{i[j]}$ is necessarily smaller than 1,
implying that this provides a tighter bounds on $x_{i[j]}$.
\par
To use this approximation to compute better bounds than the ones given by simple
interval arithmetic, it is possible to leverage the feed-forward
structure of the neural networks and obtain bounds one layer at a time. Having
included all the constraints up until the $i$-th layer (not including the $i$-th layer), it is possible to
optimize over the resulting linear programming problem and obtain bounds for all the units
of the $i$-th layer, which in turn will allow us to create the
constraints Equation~\ref{eq:planet-relu-hull} for the next layer.
\par
In addition to the pruning obtained by the convex relaxation, both Planet and
Reluplex make use of conflict analysis~\citep{Marques-Silva1999} to discover
combinations of splits that cannot lead to satisfiable assignments, allowing
them to perform further pruning of (sub-)problems.
\par
\textbf{Toy Example (Running Planet)}
\textit{Planet first computes initial bounds via interval arithmetic for nodes a, b and y. Then it builds a linear program to approximate the network by using the linear constraints Equations~\ref{eq:planet-reluout-lbinp}, \ref{eq:planet-reluout-lb0} and \ref{eq:planet-reluout-ub}. Upper and lower bounds of a, b and y are refined by calling an LP solver. In this toy example, after refinement, the lower bound and upper bound of y are replaced with $-3$ and $5$ respectively. Since it is sufficient to prove the property with the refined lower bound, the algorithm exits before entering the main loop where a Satisifiability solver is called. }

\section{Improved BaB for NN Verification }
As can be seen, previous approaches to neural network verification have relied
on methodologies developed in three communities: optimization, for the estimation
of upper and lower bounds; verification, especially SMT; and machine learning,
especially the feed-forward nature of neural networks for the creation of
relaxations. A natural question that arises is ``Can other existing literature
from these domains be exploited to further improve neural network
verification?''  Our Branch-and-Bound framework makes it easy to answer
this question. With its help, we can easily identify and consequently provide a non-exhaustive list of techniques to speed-up verification algorithms.

\subsection{Better Bounding} 
\label{subsec:bounds}
While the relaxation proposed by \citet{Ehlers2017} is tighter than the one used by Reluplex, it can be improved further still. Specifically, after a splitting operation, on a newly generated (sub-)problem,
we can refine all the $\mbf{l_i}, \mbf{u_i}$ bounds by, for instance, formulating corresponding linear programming problems with added constraints from the split. Then, we solve for minimum for $l_{i[j]}$ and maximum for $u_{i[j]}$. With refined upper and lower bounds, we are able to introduce smaller convex relaxation hulls to obtain tighter
relaxations. We show the importance of this in the experiments section with the
\textbf{reluBaB} method that performs splitting on the activation like
Planet but updates its intermediate bounds approximation completely at each step. However, it should be noted there is a trade-off between the benefits of tighter relaxation and the overall computational efficiency. Since updating all the $\mbf{l_i}, \mbf{u_i}$ bounds could be computationally expensive, we also show in the experiments section that, depending on the problem at hand, it is sometimes sufficient to update bounds for some of the layers. The overall gain from a tighter relaxation is not in a linear relationship with the number of intermediate bounds updated.

\subsection{Better Branching}
In the following, we discuss two possible ways to improve branching strategies. 
\subsubsection{Branching on Input Domains}
In the previous discussions, both Planet and Reluplex adopt the decision to split on the activation of the ReLU non-linearities. Although the decision is intuitive as it provides a clear set of decision variables to fix, these methods ignore another natural branching strategy, namely, splitting on the input domain. There are two main advantages of input domain splitting strategies. Firstly, it is simple and straightforward to apply. Once an input dimension to split on is decided, we only need to modify the associated input constraints for each sub-domain, generated by the split step of the BaB algorithm. There is no need to deal with potential conflicts (e.g. infeasible sub-problem) that could be introduced by fixing a ReLU node. Secondly, it can be argued that since the function encoded by the neural networks are piecewise linear in their input, splitting the input domain could result in the computation of high quality upper and lower bounds. With tighter input bounds, tighter intermediate bounds at all layers can be easily re-evaluated, which might not be the case for splitting on a ReLU node, at least for layers prior to the ReLU node we branch on.
\par
To demonstrate the benefits of input domain splitting, we propose two novel input split algorithms. We will show in experiments sections that domain splitting strategies incorporating our proposed heuristics are highly effective for small scale verification problems with low dimensional input. The first and the most direct algorithm is 
\textbf{BaB} algorithm. Based on a domain with input constrained by
Equation~\ref{eq:ctx-bounds0}, the $\mtt{split}$ function would return two
subdomains where bounds would be identical in all dimension except for the
dimension with the largest length, denoted $i^\star$. The bounds for each
subdomain for dimension $i^\star$ are given by \mbox{$l_{0[i^\star]} \leq
  x_{0[i^\star]} \leq \frac{l_{0[i^\star]} + u_{0[i^\star]}}{2}$} and
\mbox{$\frac{l_{0[i^\star]} + u_{0[i^\star]}}{2} \leq x_{0[i^\star]} \leq
  u_{0[i^\star]}$}.
\par
In order to exploit the benefits of input domain splitting to the fullest, we introduce
the second splitting heuristic by using the highly efficient lower bound computation approach of \citet{Wong2018}. This approach
was initially proposed in the context of robust optimization. However, our
unified framework opens the door for its use in verification. 
We propose a smart branching method \textbf{BaBSB} to replace the longest edge
heuristic of \textbf{BaB}. We proceed as follows. For each input dimension $i$, we split on it and generate two subdomains, denoted by $sub_{i_0}$ and $sub_{i_1}$. Then, by using the fast approach in \citet{Wong2018}, we are able to compute lower bound estimations $f^{n}_{i_0}$ and $f^{n}_{i_1}$ of subdomains $sub_{i_0}$ and $sub_{i_1}$ respectively \footnote{For clarity, $f^{n}_{i_0}$ is a lower bound estimation of the minimum output of the network \textbf{net} can reach on $sub_{i_0}$}. Finally, we make the split decision by choosing the dimension that improves the domain's lower bound the most after the split, which is the solution of $\arg\max_{i}(\min(f^{n}_{i_0}, f^{n}_{i_1}))$. Here, we have used the minimum of $f^{n}_{i_0}, f^{n}_{i_1}$ to represent the improvement achieved over the split on a input dimension. It is possible to replace it with other criteria such as the $\max(f^{n}_{i_0}, f^{n}_{i_1})$ or the product of $f^{n}_{i_0}$ and $f^{n}_{i_1}$, as used in \citet{khalil2016learning} in the context of other MIP problems.
\par
In terms of computing a lower bound $f^{n}$ for a sub-domain, we assume the verification property has been reformulated and added as final layers to the network that we need to verify on. The modified network has a scalar output. This should be easily doable as discussed in section 3.1. Then, a rough estimate of a lower bound can be obtained by using the following formula introduced in \citet{Wong2018}: assume an arbitrary sub-domain is upper bounded by the vector $\mbf{u}_{0}$ and lower bounded by the vector $\mbf{l}_{0}$,
  \begin{equation}\label{eq:kw}
  \begin{split}
    f^{n} =& -\sum_{k=1}^{n} \nu_{k+1}^T b_k - \mbf{u}_{0}^T[\hat{\nu}_1]_+ + \mbf{l}_{0}^T[\hat{\nu}_1]_- \\
    & + \sum_{k=2}^{n}\sum_{j \in \mathcal{I}_i} \frac{u_{k[j]} l_{k[j]}}{u_{k[j]} - l_{k[j]}}[\hat{\nu}_{k[j]}]_+ 
  \end{split}
\end{equation}
where
\begin{align*}
    \nu_{n+1} & = -1 \\
    \hat{\nu}_k & = W_k^T \nu_{k+1}, k=n,\ldots,1 \\
    \nu_{k,j} & = \left \{
      \begin{array}{lll}
        0 & \text{if } u_{k[j]} > 0 \quad &(\, j\in \mathcal{I}^{-}_k) \\
        \hat{\nu}_{k[j]} &  \text{if }l_{k[j]} > 0 \quad  &(\, j\in \mathcal{I}^{+}_k)  \\
        \frac{u_{k[j]}}{u_{k[j]} - l_{k[j]}} [\hat{\nu}_{k[j]}]_+ - \frac{u_{k[j]}}{u_{k[j]} - l_{k[j]}} [\hat{\nu}_{k[j]}]_- & \text{otherwise }\quad &(\, j\in \mathcal{I}_k) \\
      \end{array} \right .\\
      &\kern 1.5in \for k= n, \dots, 2.   
  \end{align*}

Here, $[v]_{-}, [v]_{+}$ represent negative and positive parts of an element $v$ respectively. The term $I_k^{-}$ denotes the set of activation nodes whose upper bounds are negative while $I_k^{+}$ contains activation nodes whose lower bounds are positive. The rest of activation nodes belong to $I_k$. This approach is computationally efficient as one computation of $f^{n}$ is equivalent to one backward pass due the recursive nature of $v_{k}$ and $\hat{v}_k$. Here, we have directly used all intermediate bounds $u_{k[j]}$ and $l_{k[j]}$. These intermediate bounds are actually computed beforehand in a similar fashion with above formula. Each $l_{k[j]}$ can be treated as a lower bound to a sub-network consisting of layers prior to it and $u_{k[j]}$ are obtained by negating the signs of the sub-network. Given the input constraints, we proceed layer by layer to compute all intermediate bounds $u_{k[j]}$ and $l_{k[j]}$. 
\par
Once a domain splitting decision is made, two sub-domains are generated. On each sub-domain, we use an LP solver to refine intermediate bounds, as mentioned in Section~\ref{subsec:bounds}. Finally, we call the LP solver again to compute the domain lower bound ($\mtt{prob\_ub}$). Performance of \textbf{BaB} and \textbf{BaBSB} are included in the experiments section.

\subsubsection{Branching on ReLU Activation Nodes}
 
Despite their success in small-scale verification problems, input domain splitting methods are often found to be inadequate for large scale networks, as there are several limitations of input domain branching strategies. Firstly, for some methods (e.g. \textbf{BaBSB}), their computation cost for making a branching decision increases at least linearly with input dimensions. High computational cost renders these method infeasible for high input dimensional problems. Secondly and more importantly, potential input branching decisions constitute a fairly small portion of all potential branching decisions for large-scale network architecture, which contains a sizable number of unfixed ReLU activation nodes (each is a valid potential branching point). Focusing on input domain splitting alone significantly limits the power of verification methods. Given the wide and almost dominant usage of deep convolutional network in various tasks, developing an effective and computationally cheap heuristic for branching on ReLU non-linearities is of considerable importance for verification. 

In this section, we propose a new smart branching algorithm \textbf{BaBSR} on the activation of the ReLU non-linearities. So far, to the best of our knowledge, existing ReLU node splitting methods are \textbf{Reluplex}, \textbf{Planet} and \textbf{Neurify}. We show that \textbf{BaBSR} enjoys various benefits over the existing methods, although all methods use the same \textbf{branching rule}: for a given ReLU node (a node \mbox{$x_{i[j]} =\max\left( \xhat_{i[j]}, 0 \right)$} is split out into two subdomains:\mbox{$\{x_{i[j]}=0, \xhat_{i[j]}\leq 0\}$} and \mbox{$\{x_{i[j]}=\xhat_{i[j]},\xhat_{i[j]}\geq 0\}$}). To start, unlike \textbf{Planet} which does not have any heuristic to make splitting decisions, \textbf{BaBSR} uses a simple and fast heuristic to prioritise which unfixed ReLU node to split on. \textbf{Reluplex} uses the SMT core to handle the splitting order and \textbf{Neurify} computes gradient scores to prioritise ReLU nodes. We show in experiments that the prioritising strategy of \textbf{BaBSR} is much more successful than that of \textbf{Reluplex} and \textbf{Neurify} in selecting an effective ReLU node. 
Additionally, \textbf{BaBSR} has a convergence guarantee and encourages early termination, which means verification problems can be solved completely and efficiently. Once a ReLU node has been selected, \textbf{BaBSR} calls a commercial solver to obtain a lower bound for each sub-problem with the added constraint $ \xhat_{i[j]}\leq 0$ or $ \xhat_{i[j]}\geq 0$ respectively. The algorithm continues until line 5 in Algorithm 1 is satisfied. Convergence guarantee is inherently supported by the algorithm while early termination through finding adversarial examples is assisted by the prioritisation we used in generating sub-problems. 
  \par
  The heuristic used for prioritising ReLU nodes is based on the similar idea to the one used in \textbf{BaBSB}. For an arbitrary picked-out domain, we refer to the lower bound of the network minimum on the domain as $f^{n}$. In order to decide which unfixed ReLU nodes to split on, for each potential splitting option (any unfixed ReLU node), we attempt to compute a rough estimate of the potential improvement 
  to the lower bound. We then make the splitting decision by choosing the unfixed node with the largest estimated improvement.
  \par
  In detail, we estimate the improvement on splitting an arbitrary unfixed ReLU node via a modified application of Equation~\ref{eq:kw}. We first observe that when imposing a constraint on an arbitrary unfixed ReLU node $x_{i[j]}$, we will force $\nu_{i[j]}$ to be either $0$ (ReLU is in a blocking state) or $\hat{\nu}_{i[j]}$ (ReLU is in a completely passing state). As a direct result, the associated terms $\nu_{i[j]}b_{i-1[j]}$ and $\frac{u_{i[j]}l_{i[j]}}{u_{i[j]} - l_{i[j]}} [\hat{\nu}_{i[j]}]_+$ in Equation~\ref{eq:kw} will change accordingly. Furthermore, $\nu_k$, $\hat{\nu}_{k}$ for $k<i$ as products of $\nu_{i}$ will take different values as well and so do their associated terms in Equation~\ref{eq:kw} . It is possible to compute lower bounds $f^{n}$ by assuming $\nu_{i[j]} = 0$ and $\nu_{i[j]} =\hat{\nu}_{i[j]}$ and then take the minimum (or maximum, product etc.) of the two cases to represent the improvement made if splitting on $x_{i[j]}$. However, doing this would require two full or partial (only $\nu_k$, $\hat{\nu}_{k}$ for $k<i$ need to be updated) backward passes for one ReLU splitting choice, which can be computationally expensive when the number of unfixed ReLU nodes is large. To deal with this issue, we use a key observation when dealing with different data sets: when the weights (similar for bias) on each layer are of same magnitudes, the potential improvement of each $x_{i[j]}$ is generally dominated by the changes of terms $\nu_{i[j]}b_{i-1[j]}$ and $\frac{u_{i[j]}l_{i[j]}}{u_{i[j]} - l_{i[j]}} [\hat{\nu}_{i[j]}]_+$. Since a rough estimation is sufficient in this scenario, we thus propose to evaluate each ReLU split choice $x_{i[j]}$ by computing a ReLU score $s_{i[j]}$, defined as
  \begin{equation}\label{eq:kw_score}
  s_{i[j]}=\left|\min\left(\frac{u_{i[j]}}{u_{i[j]} - l_{i[j]}}\hat{\nu}_{i[j]}b_{i-1[j]},\: (\frac{u_{i[j]}}{u_{i[j]} - l_{i[j]}}-1)\hat{\nu}_{i[j]}b_{i-1[j]}\right) - \frac{u_{i[j]}l_{i[j]}}{u_{i[j]} - l_{i[j]}} [\hat{\nu}_{i[j]}]_+\right|.
  \end{equation}
 We make a decision by picking the ReLU with the largest score. One major benefit of this heuristics is that it is highly computationally efficient. As the same $v_{i}$ and $\hat{v}_{i}$, for $1<i<n-1$, are used for computing all unfixed ReLUs scores, the recursive formulations of $v_{i}$ and $\hat{v}_{i}$ enable us to compute all ReLU scores within one single backward pass, regardless the total number of unfixed ReLU nodes.   
 \par
 To maximize the performance of the heuristic, we have also incorporated into the heuristic other useful observations. Firstly, we refer to a convolution layer as sparse if it contains mostly zeros when linearlized. We found that splitting on the sparsest layer is often ineffective in terms of improving the global lower bound when compared with choices on other layers. In addition, ReLU scores are likely to fail in giving good indications when all of them are close to zero. Thus, given a network that contains a large convolution layer with a small kernel, we do not consider the unfixed ReLU nodes on this layer until all other improvements computed are relatively small. When this happens, we consider the heuristic used in prioritising ReLU nodes to be no longer effective. Hence, other selecting strategies should be used, such as, random selections with a preference over non-sparse layers.
  \par 
 Finally, unlike \textbf{BaBSB}, \textbf{BaBSR} does not call an LP solver to compute tight intermediate bounds. The main applications of \textbf{BaBSR} should be networks with large convolution layers. For these networks, computing tight intermediate bounds via an LP solver is computationally infeasible. Thus, once a ReLU split choice is made, we explicitly replace the upper or lower bound of the corresponding ReLU node to 0 for each newly generated sub-problem and then update intermediate bounds by the better of interval arithmetic bounds and bounds computed using the method of \citet{Wong2018}. Overall, with rough improvement estimations of ReLU choices and loose intermediate bounds, \textbf{BaBSR} is significantly cheap for each branching step. While many more branches might be required to solve a property, we often find this trade-off is worthwhile as will be seen in the experiments section. 

\subsection{Other Potential Improvements}
We also list some potential improvements that could be made in future research. One possible area of improvement lies in the tightness of the bounds used. We note that Equation~\ref{eq:planet-relu-hull} is very closely related to the Mixed
Integer Formulation of Equation~\ref{eq:mip-form0}. In fact, it corresponds to
level 0 of the Sherali-Adams hierarchy of relaxations~\citep{Sherali1994}. One possible improvement is to use stronger
relaxations by exploring higher levels of the hierarchy. This
would jointly constrain groups of ReLUs, rather than linearising them
independently. A related work is that of \citet{Anderson2019}, in which a MIP formulation for neural networks using stronger relaxations is proposed.
\par
One advantage of the Branch-and-Bound framework is that it is not restricted to
piecewise linear networks, which is not true for methods such as \textbf{Reluplex},
\textbf{Planet} or the MIP encoding. Any type
of networks for which an appropriate bounding function can be found will be
verifiable with a Branch-and-Bound based method. In order for Branch-and-Bound to achieve good performance on various kinds of networks, developing appropriate bounding functions is necessary. Recent advances on bounds for activations such as sigmoid or hyperbolic tangent have been made in \citet{Dvijotham2018}. While their focus is on incomplete methods, our Branch-and-Bound framework makes it readily usable for complete verification as well.

\section{Experimental Setup}
The problem of PL-NN verification has been shown to be
NP-complete~\citep{Katz2017}. Meaningful comparison between approaches therefore
needs to be experimental. We use a timeout of two hours for each experiment, unless otherwise stated.

\subsection{Methods}
The simplest baseline we refer to is \textbf{BlackBox}, a direct encoding of
Equation~\ref{eq:sat-problem} into the Gurobi solver, which will perform its own
relaxation, without taking advantage of the problem's structure.
\par
For the SMT based methods, \textbf{Reluplex} and \textbf{Planet}, we use the
publicly available versions~\citep{planetGH,reluplexGH}. Both tools are
implemented in C++. We wrote software to support conversion between the input formats of both solvers in both directions. However, it is worth noting that, we do not change the underlying GLPK solver for linear programming. All the other methods use a potentially faster Gurobi LP solver. The reader is reminded to take this key difference into account when studying our results.
\par
We also evaluate the potential of using MIP solvers, based on the formulation of
Equation~\ref{eq:mip-form0}. Due to the lack of availability of open-sourced methods
at the time of our experiments, we reimplemented the approach in Python, using
the Gurobi MIP solver. We report results for a variant called
\textbf{MIPplanet}, which uses bounds derived from Planet's convex relaxation
rather than simple interval arithmetic. Both the \textbf{MIPplanet} and \textbf{BlackBox} are
not treated as simple feasibility problem but are encoded to minimize the
output $\xhat_n$ of Equation~\ref{eq:final0}, with a callback interrupting the
optimization as soon as a negative value is found. Additional discussions on
encodings of the MIP problem can be found in the appendix.
\par
\begin{figure}
  \centering
  \begin{minipage}[t]{.80\textwidth}
    \centering
    \begin{subfigure}[t]{.4\textwidth}
      \includegraphics[width=\textwidth]{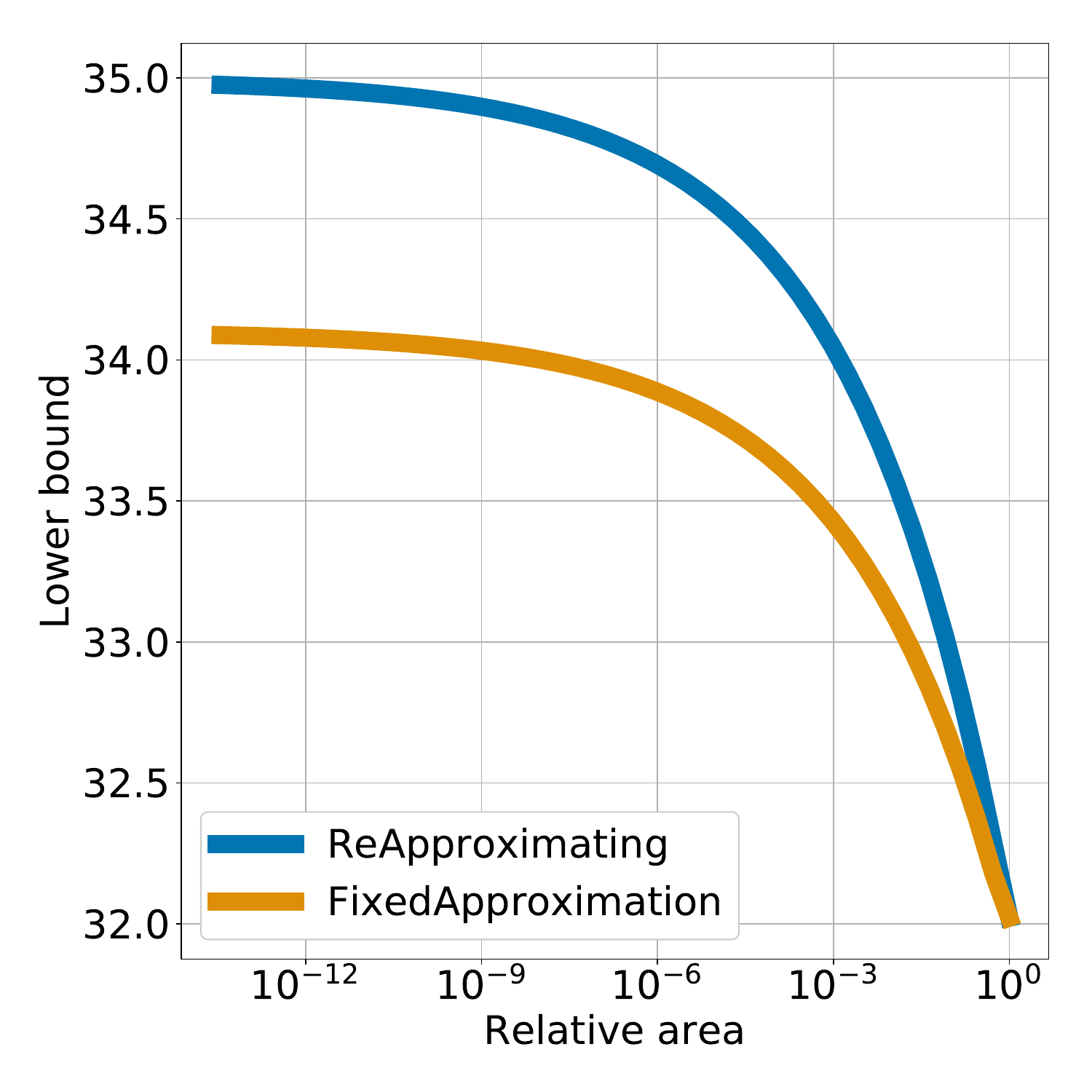}
      \caption{\label{fig:shallow-approx} Approximation on a
        \mbox{CollisionDetection} net.}
    \end{subfigure}%
    \hspace{.1\textwidth}%
    \begin{subfigure}[t]{.4\textwidth}
      \includegraphics[width=\textwidth]{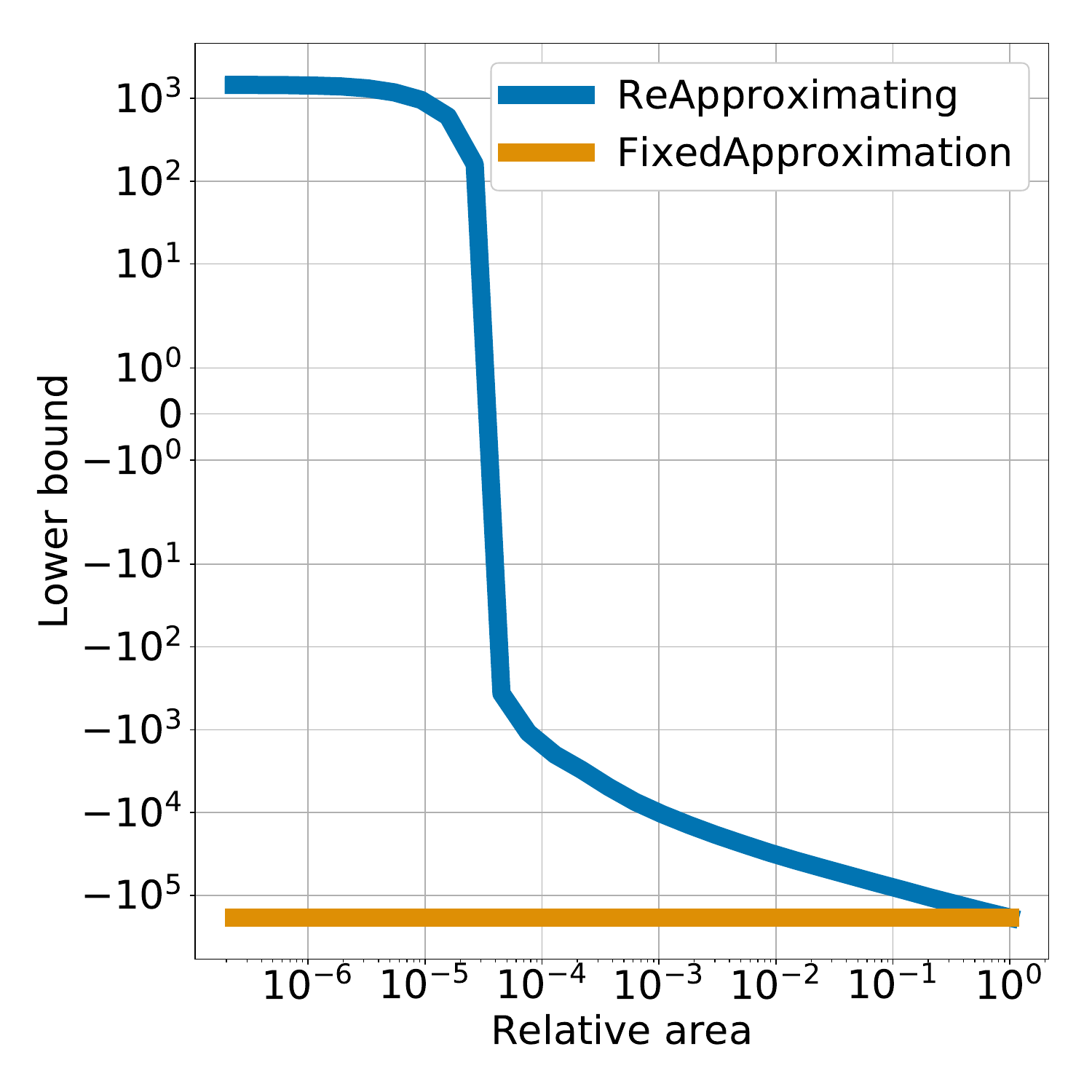}
      \caption{\label{fig:deep-approx} Approximation on a deep net from ACAS.}
    \end{subfigure}
  \end{minipage}

  \begin{minipage}[t]{\textwidth}
    \caption{\label{fig:lin-approx} Quality of the linear approximation,
        depending on the size of the input domain. We plot the value of the lower
        bound as a function of the area on which it is computed (higher is
        better). The domains are centered around the global minimum and repeatedly
        shrunk. Rebuilding completely the linear approximation at each step allows
        to create tighter lower-bounds thanks to better $\mbf{l_i}$ and
        $\mbf{u_i}$, as opposed to using the same constraints and only changing
        the bounds on input variables. This effect is even more significant on
        deeper networks.}
  \end{minipage}
\end{figure}

In our benchmark, we include the methods derived from our Branch-and-Bound
analysis. Our implementation follows Algorithm~\ref{alg:bab} faithfully, is
implemented in Python and uses Gurobi to solve LPs. The $\mtt{pick\_out}$
strategy consists in prioritising the (sub-)problem that currently has the smallest
lower bound. Upper bounds are generated by randomly sampling points on the
considered (sub-)problem or directly compute the network value at the lower bound solution provided by Gurobi, for which we use the convex approximation of \citet{Ehlers2017} to
obtain lower bounds. Motivated by the observation shown in Figure~\ref{fig:lin-approx}, which demonstrates the significant improvements it brings especially for deeper
networks, we do not always using a single approximation of the network as was done
in \citet{Ehlers2017}. Bearing in mind the trade-off between benefits of tighter relaxation and computational costs, we rebuild the approximation via calling an LP solver to recompute none or partial or all intermediate bounds for each sub-problem. This decision should be made on a case by case basis depending on the size of a network architecture, the number of the input dimensions and the magnitude and the correlation of layer weights. To better study the trade-off, we have incorporated different approximation strategies into different algorithms and compared them on several data sets.

For $\mtt{split}$, we focus on two types of split: input domain split and ReLU activation split. In each case, we consider naive split methods and improved versions. Specifically, in terms of input domain split, the naive methods (e.g. \textbf{BaB}) simply performs branching by splitting the input domain in half along its longest edge, while the improved methods (e.g. \textbf{BaBSB}) does it by
splitting the input domain along the dimension that gives the estimated best improvement to the global lower bound. Estimations are made through the fast bounds formula provided in \citet{Wong2018}. Regarding the ReLU node split, we study methods which always split on the first or a random unfixed ReLU node on the first layer containing unfixed ReLU nodes. More advanced methods (e.g. \textbf{BaBSR} and \textbf{BaBSRL}) prioritize ReLU nodes through the criteria introduced in Equation~\ref{eq:kw_score}. Each specific method is a combination of the choice of approximation strategies (how to rebuild approximations) and branching strategies (what branching heuristics to use). We summarize all Branch-and-Bound methods considered in Table~\ref{tab:all_BaB_methods}.\footnote{We mention one implementation strategy used to achieve improved performances. For all BaB methods, an LP solver is called to compute the lower bound of a subdomain. In our case, the LP solver used is Gurobi. We found that when Gurobi is used to compute all intermediate bounds, it is faster to reintroduce the LP problem in a layer by layer order to Gurobi for each sub-problem. That is, given a sub-problem, we first create a new Gurobi model instance and introduce all constraints up to the first non-activation layer. Then we compute all upper and lower bounds for the layer via Gurobi. After this is done, we add constraints up to the second non-activation layer and compute all bounds. This procedure continues until we reach the final layer and obtain a lower bound of the sub-problem by solving the corresponding LP problem in its complete form. However, for methods that do not require tight intermediate bounds, obtained via an LP solver, it is cheaper to create a single Gurobi model instance at the start. Then for each sub-problem, we only update the constraints of the Gurobi model to be consistent with the LP of the sub-problem and compute a lower bound for it.} We use abbreviations \textbf{imb} for intermediate bounds and $\mtt{prob\_ub}$ and $\mtt{prob\_lb}$ are the same as those defined in Algorithm~\ref{alg:bab}.

{
\begin{table}[h!]
  \begin{minipage}[t]{0.95\textwidth}
    \vskip 0pt
  \resizebox{\textwidth}{!}{
  \begin{tabular}{|c|c|c|l|}
    \hline 
    & & & \\
    \textbf{Method} & \textbf{Branching Type} & \textbf{Branching Heuristics}& \multicolumn{1}{c|}{\textbf{Approximations}} \\ 
   & & & \\
    \hline
    BaBSB & \multirow{2}{*}{\begin{tabular}{@{}c@{}}\\   \textbf{Input}\end{tabular}} & \begin{tabular}{@{}c@{}}fast bounds \\ \citet{Wong2018} \end{tabular} & \begin{tabular}{ll} \textbf{imb: }&  LP solver \\
    $\mtt{prob\_lb}$& LP solver\\$\mtt{prob\_ub}$& random sampling \end{tabular} \\
    \cline{1-1}\cline{3-4}
    BaB &  & longest edge &\begin{tabular}{ll} \textbf{imb: }&  LP solver \\
    $\mtt{prob\_lb}$& LP solver\\$\mtt{prob\_ub}$& random sampling \end{tabular} \\
    \hline
    reluBaB & \multirow{3}{*}{\begin{tabular}{@{}c@{}}\\ \\ \\ \\ \\ \\  \textbf{ReLU}\end{tabular}} & \begin{tabular}{c} the first unfixed ReLU node 
                                    \\ on the first layer \\containing unfixed ReLU nodes  
                                     \end{tabular} & \begin{tabular}{ll} \textbf{imb: }& LP solver \\
    $\mtt{prob\_lb}$& LP solver\\$\mtt{prob\_ub}$& random sampling \end{tabular}\\
    \cline{1-1}\cline{3-4}
    
    BaBSR & & \begin{tabular}{@{}c@{}} prioritization via the criteria \\ defined in Equation~\ref{eq:kw_score} \end{tabular} &  \begin{tabular}{ll} \textbf{imb: }& better of interval bounds \\ &and bounds of \\&\citet{Wong2018}\\
    $\mtt{prob\_lb}$& LP solver\\$\mtt{prob\_ub}$& solution of the LP \end{tabular} \\
    \cline{1-1}\cline{3-4}
    
    BaBSRL &  & \begin{tabular}{@{}c@{}} prioritization via the criteria \\ defined in Equation~\ref{eq:kw_score} \end{tabular} & \begin{tabular}{ll} \textbf{imb: }& update the bounds of the layer \\ &right before \\&the ReLU node selected \\&via an LP solver; \\ &update the rest of the layers via\\& better of interval bounds \\ &and bounds of \\&\citet{Wong2018}\\
    $\mtt{prob\_lb}$& LP solver\\$\mtt{prob\_ub}$& solution of the LP \end{tabular} \\
    \hline
  \end{tabular}}
\end{minipage}%
\hfill%
\begin{minipage}[t]{.95\textwidth}
  \caption{\label{tab:all_BaB_methods} All Branch-and-Bound methods considered in the experiments section. }
\end{minipage}
\vspace{-20pt}
\end{table}
}

\subsection{Data Sets}
We perform verification on six data sets of properties and report
the comparison results.
\par
The \textbf{CollisionDetection} data set~\citep{Ehlers2017} attempts to predict
whether two vehicles with parameterized trajectories are going to collide.
500 properties are extracted from problems arising from a binary search to identify
the size of the region around training examples in which the prediction of the
network does not change. The network used is relatively shallow but due to the
process used to generate the properties, some lie extremely close between the
decision boundary between satisfiable (SAT) and unsatisfiable (UNSAT). Recall that satisfiable refers to properties that are false
(a counterexample if found) while unsatisfiable refers to properties that are true (no counterexample exists). Results presented in
Figure~\ref{fig:cactus_collision} therefore highlight the accuracy of methods.
\par
The \textbf{Airborne Collision Avoidance System (ACAS)} data set, as released by
\citet{Katz2017} is a neural network based advisory system recommending
horizontal manoeuvres for an aircraft in order to avoid collisions, based on
sensor measurements. Each of the five possible manoeuvres is assigned a score by
the neural network and the action with the minimum score is chosen. The 188
properties to verify are based on some specification describing various
scenarios. Due to the deeper network involved, this data set is useful in
highlighting the scalability of the various algorithms.
\par
The \textbf{Robust MNIST Network} is adopted from the network trained with the strategies proposed in \citet{Wong2018}. The network contains 2 convolution layers followed by 2 fully connected layers with a total number of 4804 activation nodes. Since on a network of this size each LP requires more than 2 seconds, the total number of branches that could be taken within a timeout is low. To better evaluate the performance of the branching heuristic used in \textbf{BaBSR}, we also introduce a reduced version, \textbf{reduced Robust MNIST Network}. The reduced network has the same structure as the original one but fewer hidden nodes on each hidden layer. The total number of ReLU nodes in the reduced network is 1226. On the reduced network, each LP requires only 0.17 seconds, which allows a large number of branching decisions to be made before timeout. For MNIST networks,
the natural properties to verify are whether the predicted label changes if each input image is allowed to be perturbed within an $\epsilon$-infinity norm ball.\footnote{For a given image $x$ with the predicted label $y^{\text{targ}}$ and a given $\epsilon$, the property to be verified is $\max(f(x')_{y^{\ast}} - f(x')_{y^{\text{targ}}}) < 0$ for $\forall x'$ s.t. $\Vert x'-x\Vert_{\infty} < \epsilon$, where $y^{\ast}$ is any label. For \textbf{MIPplanet}, the encoding of the $\max$ function is given in Appendix B.1. For other methods, the $\max$ function is encoded as a combination of linear functions and ReLUs as introduced in Appendix B.2. Although it is conceptually simpler to deal with $\max$ function directly, we saw improved performance when we replace $\max$ function with ReLUs and hence the encoding decisions.} Since each combination of an image in the MNIST test set and an $\epsilon$ constitute a valid property, we randomly select test images and verify properties at a set of pre-specified epsilons ranging from 0.14 to 0.175 for \textbf{Robust MNIST Network} and from 0.11 to 0.14 for \textbf{reduced Robust MNIST Network}. Recall that, on the same network, epsilon values determine the difficulty level of verification properties. We use a set of epsilon values for both networks to allow comprehensive evaluations. Higher value of $\epsilon$ is used for the large network, as the network is more robust. Due to the large number of properties, we restrict the timeout to be one hour on these two data sets.
\par
Existing data sets do not allow us to explore the impact of various problem/model parameters such as depth, number of
hidden units, input dimensionality and correlation between hidden nodes on the same layer. Our data sets, \textbf{PCAMNIST} and \textbf{TwinStream},
remove this deficiency, and can prove helpful in analysing future verification
approaches as well. They are generated in a way to give control over different
parameters. Specifically, \textbf{PCAMNIST} is mainly used for evaluating methods over different network architecture. It has a much wider range in terms of depth, number of hidden units and input dimensionality than \textbf{TwinStream} but no particular layer correlation is introduced. On the other hand, \textbf{TwinStream} is specially designed such that hidden nodes on the same layer are highly correlated. It allows us to explore the trade-off between different bounding strategies. Details of the data set construction are given in
the appendix. 
\par
 Finally, we summarize in Table~\ref{tab:problem_size} the characteristics of all of the
data sets used for the experimental comparison.
{
\begin{table}[h!]
  \begin{minipage}[t]{.95\textwidth}
    \vskip 0pt
  \resizebox{\textwidth}{!}{
  \begin{tabular}{|c|c|c|}
    \hline
    \textbf{Data set} & \textbf{Count} & \textbf{Model Architecture}\\
    \hline
    \begin{tabular}{l}
      Collision\\
      Detection
   \end{tabular}& 500 & \begin{tabular}{@{}c@{}}6 inputs\\
                                 40 hidden unit layer, MaxPool \\
                                 19 hidden unit layer\\
                                 2 outputs\\
                               \end{tabular}\\
    \hline
    ACAS & 188 & \begin{tabular}{@{}c@{}}5 inputs\\
                   6 layers of 50 hidden units\\
                   5 outputs
                 \end{tabular} \\
    \hline
    PCAMNIST & 27 & \begin{tabular}{@{}c@{}} 10 or \{5, 10, 25, 100, 500, 784\} inputs\\
                      4 or \{2, 3, 4, 5, 6, 7\} layers \\
                      of 25 or \{10, 15, 25, 50, 100\} hidden units,\\
                      1 output, with a margin of +1000 or \\
                      \{-1e4, -1000, -100, -50, -10, -1 ,1, 10, 50, 100, 1000, 1e4\}\\
                      \end{tabular}\\
    \hline
    \begin{tabular}{l}
      reduced\\
      ROBUST\\
      MNIST
   \end{tabular} & 1200 & \begin{tabular}{@{}c@{}} 28 by 28 inputs\\
                      Conv2d(1,4,4, stride=2, padding=1) \\
                      Conv2d(4,8,4, stride=2, padding=1)\\
                      linear layer of 50 hidden units \\
                      linear layer of 10 hidden units\\
                      $\mathcal{L}_{\infty}$ ball radius selected are \\
                      \{0.11, 0.115, 0.12, 0.125, 0.127, 0.13, 0.14\}
                      \end{tabular}\\
    \hline
    \begin{tabular}{l}
      ROBUST\\
      MNIST
   \end{tabular} & 1000 & \begin{tabular}{@{}c@{}} 28 by 28 inputs\\
                      Conv2d(1,16,4, stride=2, padding=1) \\
                      Conv2d(16,32,4, stride=2, padding=1)\\
                      linear layer of 100 hidden units \\
                      linear layer of 10 hidden units\\
                      $\mathcal{L}_{\infty}$ ball radius selected are \\
                      \{0.14, 0.15, 0.155, 0.16, 0.165, 0.17, 0.175\}
                      \end{tabular}\\
    \hline
    TwinStream & 81 & \begin{tabular}{@{}c@{}} \{5, 10, 25\} inputs\\
                      \{2, 4, 5\} layers \\
                      of \{5, 10, 25\} hidden units,\\
                      1 output, with a margin of \\
                      \{1e2, 1, 10\}\\
                      \end{tabular}\\
    \hline
  \end{tabular}}
\end{minipage}%
\hfill%
\begin{minipage}[t]{.95\textwidth}
  \caption{\label{tab:problem_size} Details of all the data sets. For
    PCAMNIST, we use a base network with 10 inputs, 4 layers of 25 hidden units
    and a margin of 1000. We generate new problems by changing one
    parameter at a time, using the values inside the brackets.}
\end{minipage}
\vspace{-30pt}
\end{table}
}

\subsection{Evaluation Criteria}
For each of the data sets, we compare different methods using the same
protocol. We attempt to verify each property with a timeout of two hours 
(with an exception of one hour timeout for the reduced Robust Network experiment and the Robust Network experiment due to the large number of properties), 
and a maximum allowed memory usage of 20GB, on a single core of a machine with an
i7-5930K CPU. We measure the time taken by the solvers to either prove or
disprove the property. If the property is false and the search problem is
therefore satisfiable, we expect from the solver to exhibit a counterexample. If
the returned input is not a valid counterexample, we don't count the property as
successfully proven, even if the property is indeed satisfiable. All code and
data necessary to replicate our analysis have been released.

\section{Analysis}
We perform an ablation study to evaluate the performance of different methods on various data sets. 

\subsection{Small Networks}
We first consider small networks with low dimensional input. These are networks of CollisionDetection and ACAS. When networks are small, computing all intermediate bounds via an LP solver is computationally affordable. In these cases, gains from tighter relaxations are often significant and the tightest intermediate bounds should be used to achieve an ideal performance. As a result, methods like \textbf{BaBSR} and \textbf{BaBSRL} that employ rough estimated intermediate bounds are not included in this section.
\begin{figure}[h!]
  \centering
  \begin{minipage}[b]{\textwidth}
    \vskip 0pt
    \centering
    \begin{subfigure}[t]{.40\textwidth}
      \vskip 0pt
      \includegraphics[width=.95\textwidth]{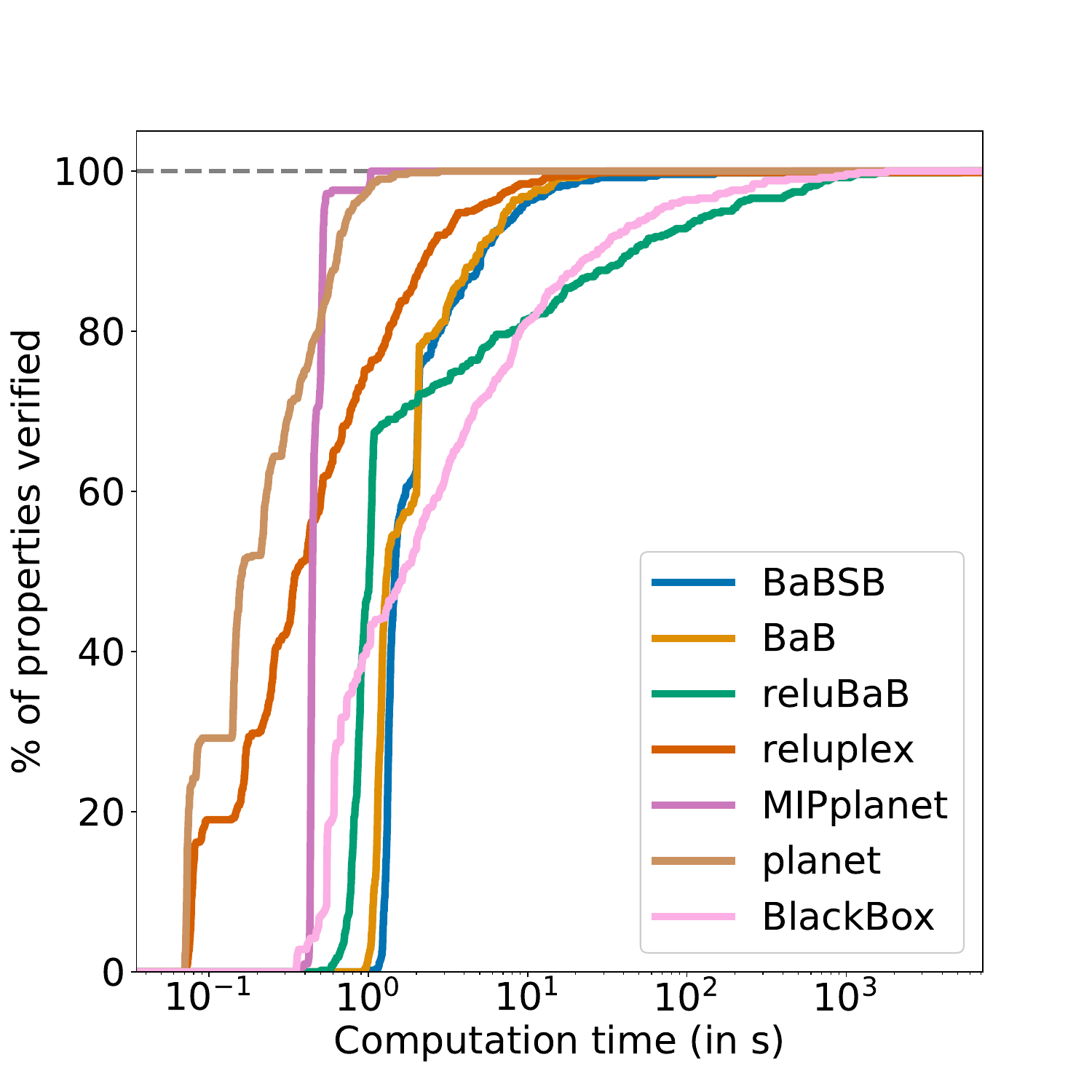}
      \caption{\label{fig:cactus_collision}CollisionDetection data set.}
    \end{subfigure}%
    \hspace{.1\textwidth}%
    \begin{subfigure}[t]{.40\textwidth}
      \vskip 0pt
      \includegraphics[width=.95\textwidth]{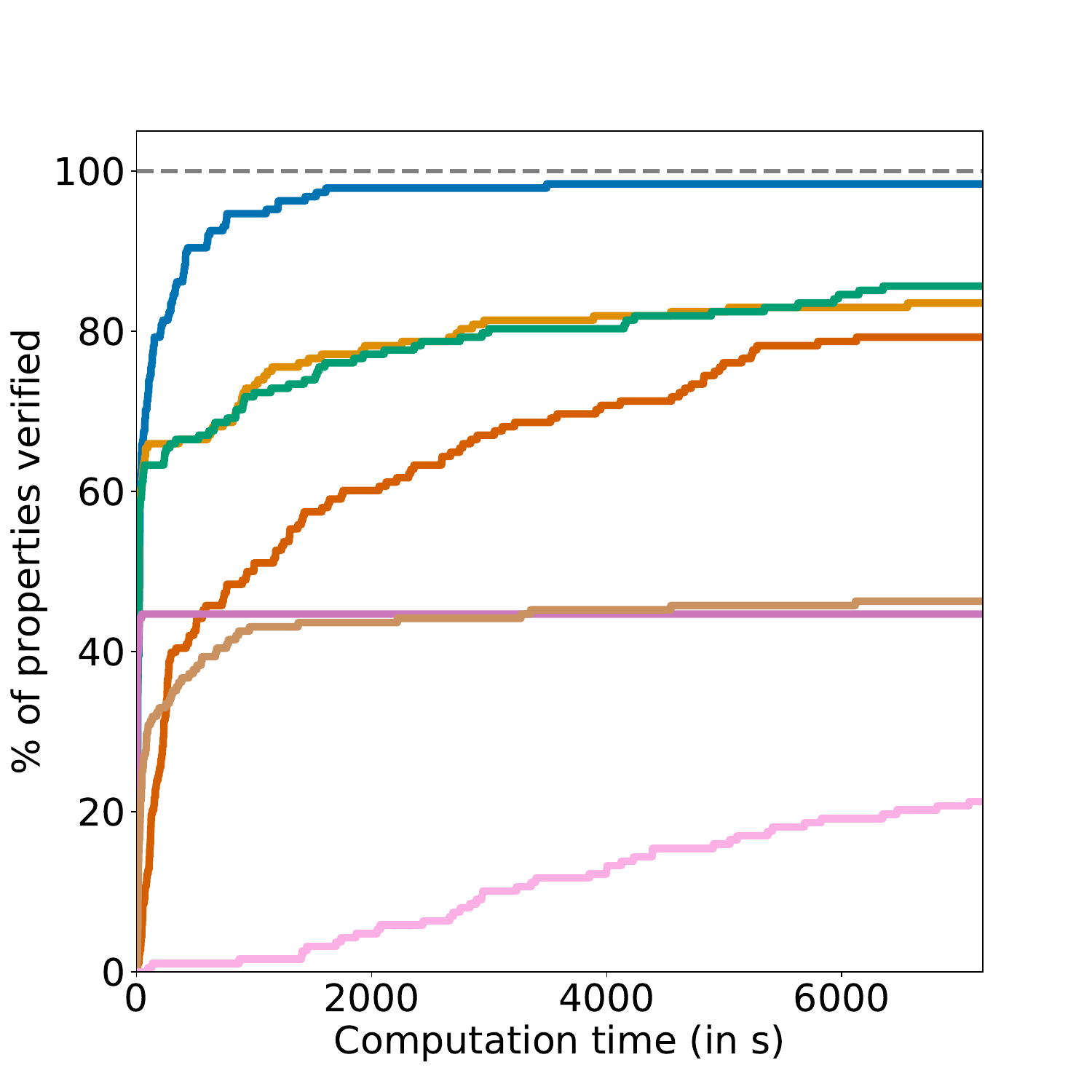}
      \caption{\label{fig:cactus_acas}ACAS data set.}
    \end{subfigure}
  \end{minipage}

  \begin{minipage}[t]{\textwidth}
    \vskip 0pt
    \caption{\label{fig:cactus_acas_collision} Proportion of properties
        verifiable for varying time budgets depending on the methods employed. A
        higher curve means that for the same time budget, more properties will
        be solvable. All methods solve CollisionDetection quite quickly except
        \textbf{reluBaB}, which is much slower and \textbf{BlackBox} which
        produces several incorrect counterexamples.}
  \end{minipage}
  \vspace{-10pt}
\end{figure}
In Figure~\ref{fig:cactus_collision}, on the shallow networks of
CollisionDetection, most solvers succeed against all properties in
about 10 seconds. In particular, the SMT inspired solvers \textbf{Planet},
\textbf{Reluplex} and the MIP solver are extremely fast. On the deeper networks of ACAS, in Figure~\ref{fig:cactus_acas}, no
errors are observed but most methods timeout on the most challenging testcases.
The best baseline is \textbf{Reluplex}, which reaches 79.26\% success rate at the two hour
timeout, while our best method, \textbf{BaBSB}, already achieves 98.40\% with a
budget of one hour. To reach Reluplex's success rate, the required
runtime is two orders of magnitude smaller.
\par
We are also able to identify the factors that allow our methods to perform well. We point out that the only difference between \textbf{BaBSB} and \textbf{BaB} is the
smart branching, which represents a significant part of the performance gap. Furthermore, on networks with low dimensional inputs, branching over the ReLU activation nodes rather than over the inputs does not contribute
much, as shown by the small difference between \textbf{BaB} and
\textbf{reluBaB}.  The rest of the performance gap can be attributed to using better bounds:
\textbf{reluBaB} significantly outperforms \textbf{planet} while using the same
branching strategy and the same convex relaxations. The improvement comes from
the benefits of rebuilding the approximation at each step shown in
Figure~\ref{fig:lin-approx}.
\par
\begin{figure}[h]
  \begin{minipage}[c]{.35\textwidth}
    \vskip 0pt
    \resizebox{.8\textwidth}{!}{
    \begin{tabular}{lc}
      \toprule
      \textbf{Method} & \begin{tabular}{c}\textbf{Average}\\ \textbf{time}\\ \textbf{per Node}\end{tabular} \\
      \midrule
      BaBSB & 1.81s \\
      BaB & 2.11s \\
      reluBaB & 1.69s \\
      \midrule
      reluplex & 0.30s \\
      \midrule
      MIPplanet & 0.017s \\
      \midrule
      planet & 1.5e-3s \\
      \bottomrule
    \end{tabular}}
    \captionof{table}{\label{tab:nb_node_average} Average time to explore a node for
      each method.}
  \end{minipage}
  \hfill
  \begin{minipage}[c]{.6\textwidth}
    \begin{subfigure}{.70\textwidth}
      \includegraphics[width=.95\textwidth, right]{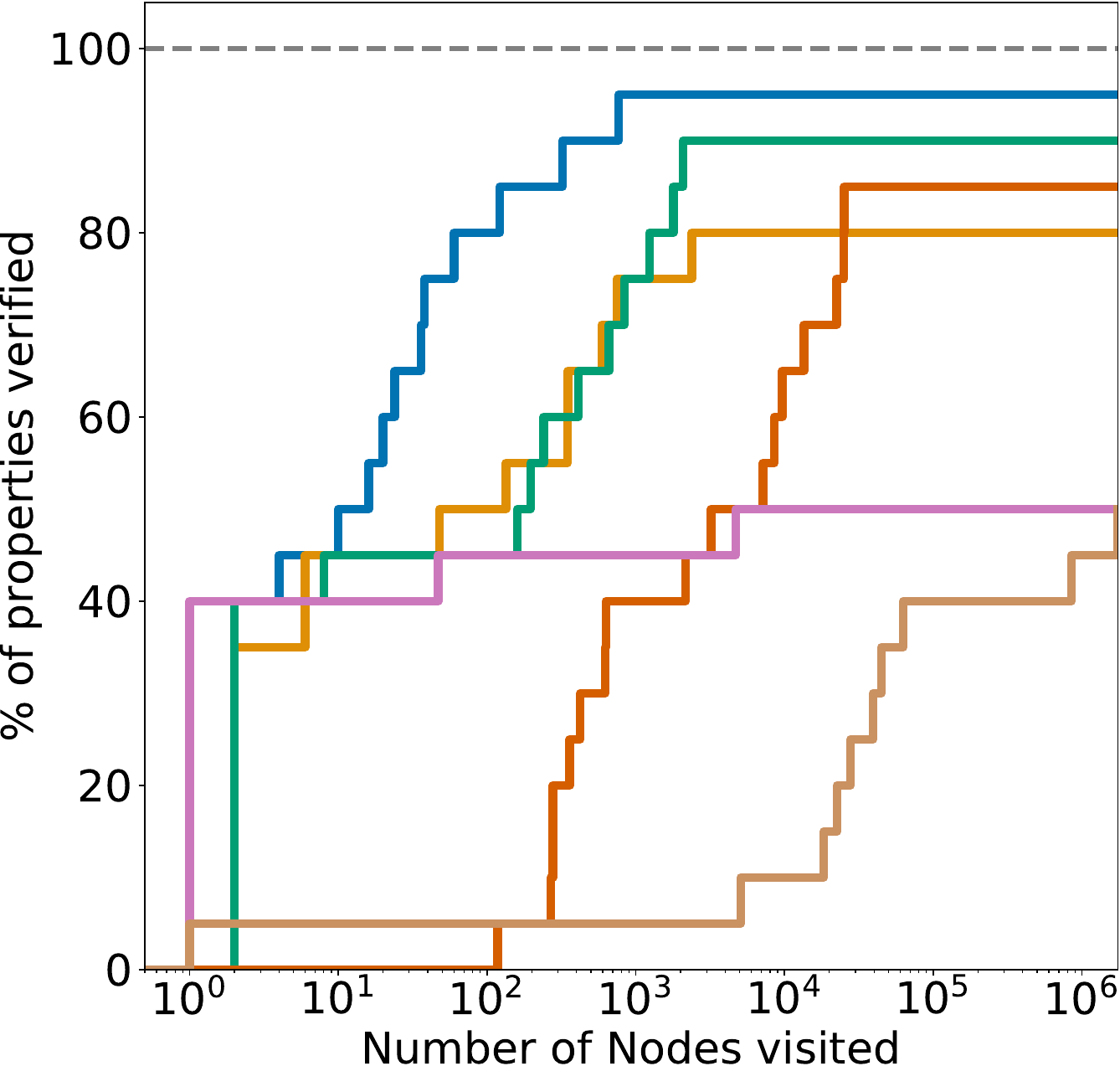}
    \end{subfigure}%
    \begin{subfigure}{.25\textwidth}
    \centering
      \includegraphics[width=0.95\textwidth, left]{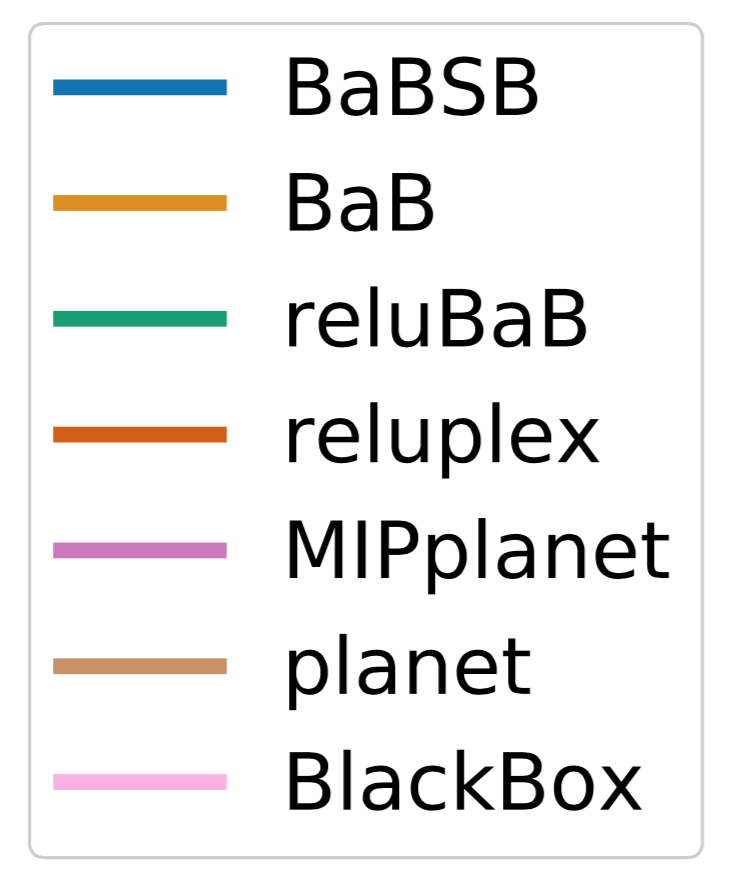}
    \end{subfigure}
    \begin{subfigure}{\textwidth}
    \caption{\label{fig:cactus_nbnode}Properties solved for a given number of
        nodes to explore (log scale).}
    \end{subfigure}
  \end{minipage}
   \caption{\label{fig:nb_node_analysis} Trade-off between bounding cost and total number of branches required. 
        Figure~\ref{fig:cactus_nbnode} shows how
        many subdomains needs to be explored before verifying properties while
        Table~\ref{tab:nb_node_average} shows the average time cost of exploring
        each subdomain. Our methods have a higher cost per node but they require
        significantly less branching, thanks to better bounding. Note also that
        between \textbf{BaBSB} and \textbf{BaB}, the smart branching reduces by
        an order of magnitude the number of nodes to visit.}

\end{figure}

Figure~\ref{fig:nb_node_analysis} presents additional analysis on a
20-property subset of the ACAS data set, showing how the methods used impact the
need for branching. Smart branching and the use of better lower bounds reduce
heavily the number of subdomains to explore.

\subsection{Large Networks}
\begin{figure}[h]
  \begin{minipage}[c]{.5\textwidth}
    \includegraphics[width=.98\textwidth]{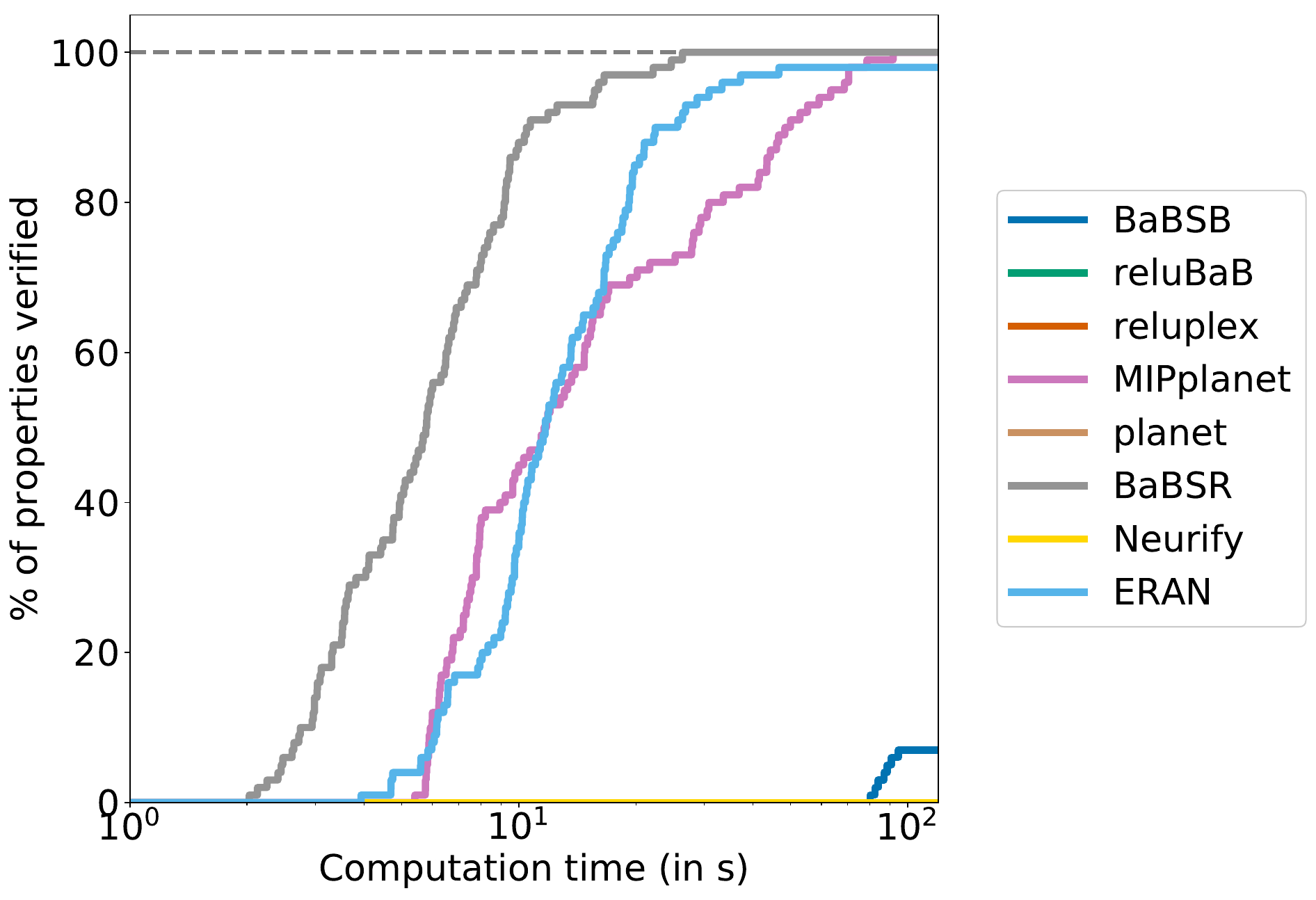}
  \end{minipage}
  \hfill
  \begin{minipage}[c]{.48\textwidth}
    \caption{\label{fig:cactus_guro_mini_time} On simple properties, all \textbf{planet}, \textbf{reluBaB}, \textbf{reluplex} and \textbf{Neurify} timed out on every single property for 100s time limit. \textbf{BaBSB} managed to solve some properties but its performance is significantly worse than that of \textbf{MIPplanet}. Only \textbf{BaBSR} and \textbf{ERAN} perform better than and similarly to \textbf{MIPplanet} and are thus ran on all properties of the reduced Robust MNIST Network and Robust MNIST Network.}
  \end{minipage}
  \vspace{-20pt}
\end{figure}

We then study the performance of various methods on the reduced Robust MNIST Network and the Robust MNIST Network. Due to the large number of properties (1200 and 1000 respectively in order to cover a wide range of difficulties), we treat \textbf{MIPplanet}\footnote{For large networks, calling an LP solver to compute all intermediate bounds in computationally expensive. To ensure a fair comparison between \textbf{BaBSR} and \textbf{MIPplanet}, the same intermediate bounds as that of \textbf{BaBSR} are used for building the LP model.} as the benchmark and rule out methods which could not perform at least at similar level to \textbf{MIPplanet} over simple properties, that is, those that could be solved within 100 seconds by \textbf{MIPplanet}. For a comprehensive study of available complete methods, we have also included \textbf{ERAN}, ETH Robustness Analyser for neural networks. It is developed on a series of work \citep{deepz2018, deeppoly2019, refinepoly2019} that apply abstract interpretation for neural network verification. \textbf{ERAN}\footnote{For our experiments, we have used the complete version of \textbf{ERAN} with refinepoly domain and $10$ seconds MILP timeout, as suggested in \citet{refinepoly2019}. All the rest of hyper-parameters are set as default values. Each process is restricted to a single cpu core.} mainly focuses on incomplete verification of MNIST, CIFAR10 and a subset of ACAS properties but also supports a complete mode, rendering itself a fair candidate for comparison studies. We observe from Figure~\ref{fig:cactus_guro_mini_time} that most methods fail even on simple properties. \textbf{BaBSB} becomes incompetent when the input dimension is high. Although \textbf{planet}, \textbf{reluplex} and \textbf{BaBSR} use the same branching rule and loose bounds of different extent for approximation, the significantly improved performance of \textbf{BaBSR} is attributed to its effective ReLU prioritization heuristic and early termination feature. The \textbf{reluBaB} method is the only ReLU split method that uses the tightest bounds available throughout the procedure. However, the fact it could not solve a single property reemphasizes the issue of finding a balance between the computational cost and the quality of the relaxation introduced. With limited computing resources, more gains might be achieved via a good branching strategy than a tighter relaxation. 

We thus only compare among \textbf{BaBSR}, \textbf{ERAN} and \textbf{MIPplanet} on all properties of reduced Robust MNIST Network and the Robust MNIST Network. Overall, \textbf{BaBSR} achieved the best performance on both data sets. In detail, when the reduced Robust MNIST network is considered, \textbf{BaBSR} outperforms \textbf{MIPplanet} significantly on easy properties, but the performance gap decreases when properties become more and more difficult. On the most challenging ones, \textbf{MIPplanet} slightly wins over \textbf{BaBSR}. A likely cause for the declined performance of \textbf{BaBSR} could be the split heuristic used. After a certain number of branches, the rough estimations used by the split heuristic are no longer fair representations of potential improvements that could be made by available branching decisions. This conjecture is consistent with what we observe on the Robust MNIST Network, where \textbf{BaBSR} significantly outperforms \textbf{MIPplanet} on all properties. Since each LP is expensive (requires more than 5 seconds) on the large network, the total number of branches that could be taken within the time limit is at least 10 times smaller than that on the reduced network, which means the heuristic of \textbf{BaBSR} probably remains effective throughout the verification procedure. \textbf{ERAN} is not as competent as the other two methods on the reduced Robust data set but it beats \textbf{MIPplanet} on the Robust data set. Compared to \textbf{MIPplanet}, \textbf{ERAN}, in its complete mode, adopts a similar idea of solving a MIP instance. However, different convex relaxations and intermediate bounds are used. Specifically, in these experiments, \textbf{ERAN} uses intermediate bounds collected by running RefinePoly analysis. Those computationally more expensive but potentially tighter intermediate bounds used by \textbf{ERAN} might explain its varying performance to that of \textbf{MIPplanet} on these data sets. In addition, when the network size increases, the time required by the LP solver increased exponentially, which becomes the main bottleneck in verifying properties on large networks in the Branch-and-Bound framework. Thus, in order to tackle real life verification problems, which often involve networks considerably larger than the Robust MNIST Network, it is important to develop an efficient LP solver that, by exploiting the special structure of neural networks, scales well with their size. At the same time, developing a better split strategy that is computationally cheap yet capable of giving high quality decisions throughout the whole Branch-and-Bound process is also key to the success of Branch-and-Bound methods on dealing with large network verification problems. One recent work~\citep{gnnBaB2020}, which employs graph neural networks to imitate strong branching decisions, has demonstrate some success in achieving desired branching strategies.  

\begin{figure}[h!]
  \centering
  \vspace{-10pt}
  \begin{minipage}[b]{\textwidth}
    \vskip 0pt
    \centering
    \begin{subfigure}[t]{.42\textwidth}
      \vskip 0pt
      \includegraphics[width=.95\textwidth]{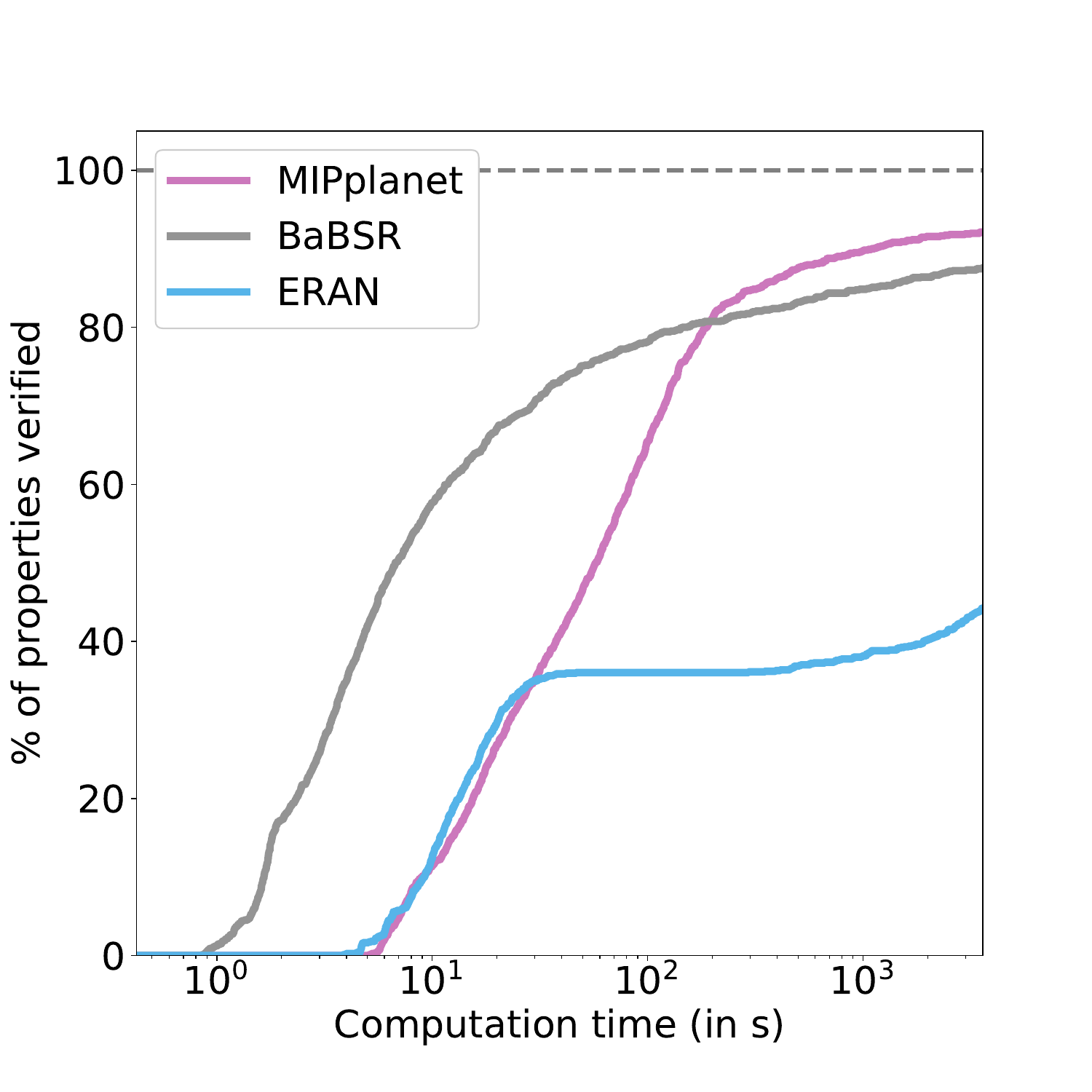}
      \caption{\label{fig:cactus_med_time}All properties of reduced Robust MNIST Network.}
    \end{subfigure}%
    \hspace{.1\textwidth}%
    \begin{subfigure}[t]{.42\textwidth}
      \vskip 0pt
      \includegraphics[width=.95\textwidth]{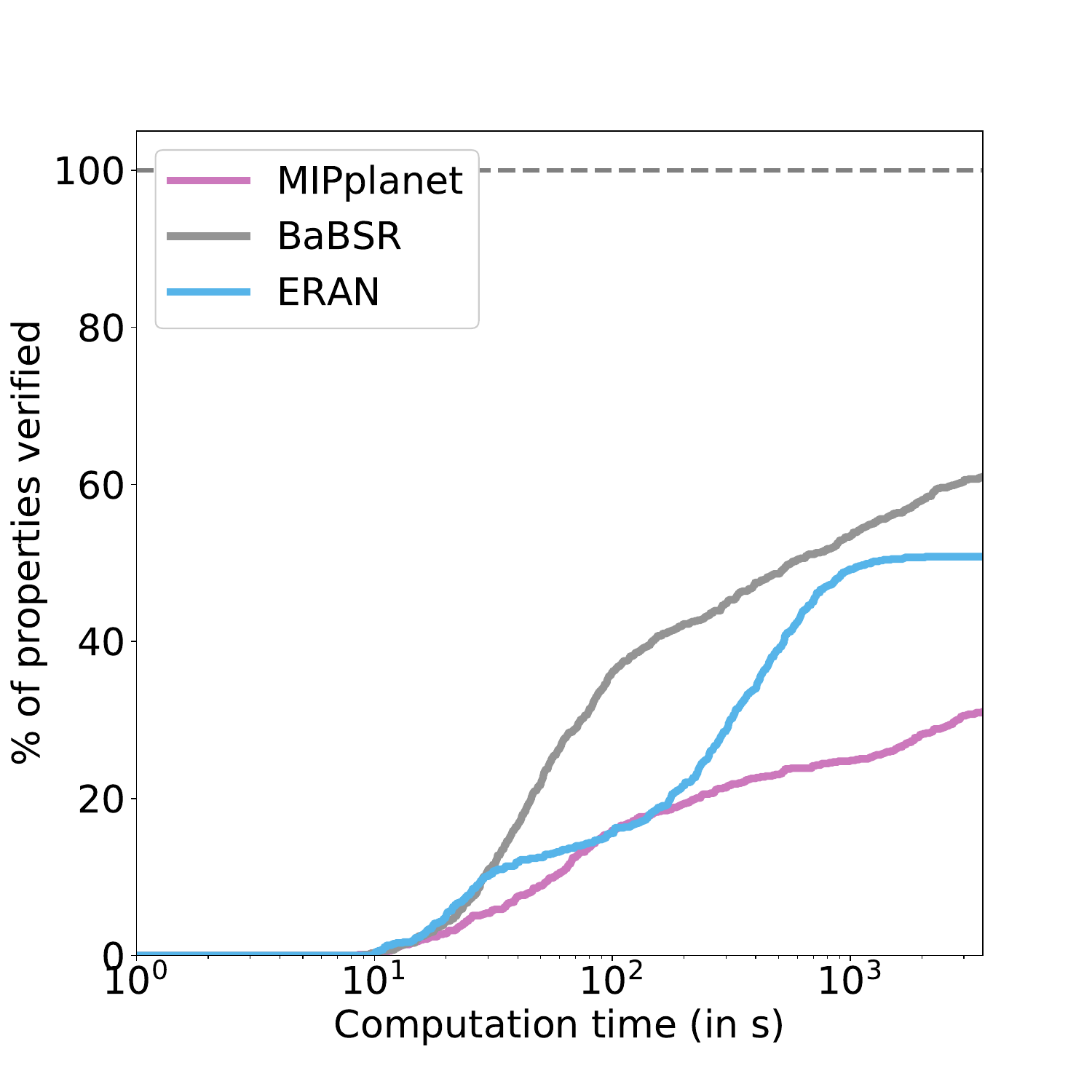}
      \caption{\label{fig:cactus_large_time}All properties of the Robust MNIST Network.}
    \end{subfigure}
  \end{minipage}

  \begin{minipage}[t]{\textwidth}
    \vskip 0pt
    \caption{\label{fig:cactn} Cactus plots of properties solved by \textbf{MIPplanet}, \textbf{BaBSR} and \textbf{ERAN} on the reduced Robust MNIST Network (left) and the Robust MNIST Network (right). \textbf{BaBSR} outperforms \textbf{MIPplanet} except on the difficult properties of the reduced Robust MNIST Network. The declined performance of \textbf{BaBSR} might be caused by the ineffectiveness of the branching heuristics at later stage. The better performance of \textbf{ERAN} than that of \textbf{MIPplanet} on the Robust MNIST Network could be explained by the possible tighter intermediate bounds used by \textbf{ERAN}.}
  \end{minipage}
  \vspace{-30pt}
\end{figure}

\subsection{Varying Parameters}
\begin{figure}
  \vspace{-5pt}
  \centering
  \begin{subfigure}{.35\textwidth}
    \includegraphics[width=\textwidth]{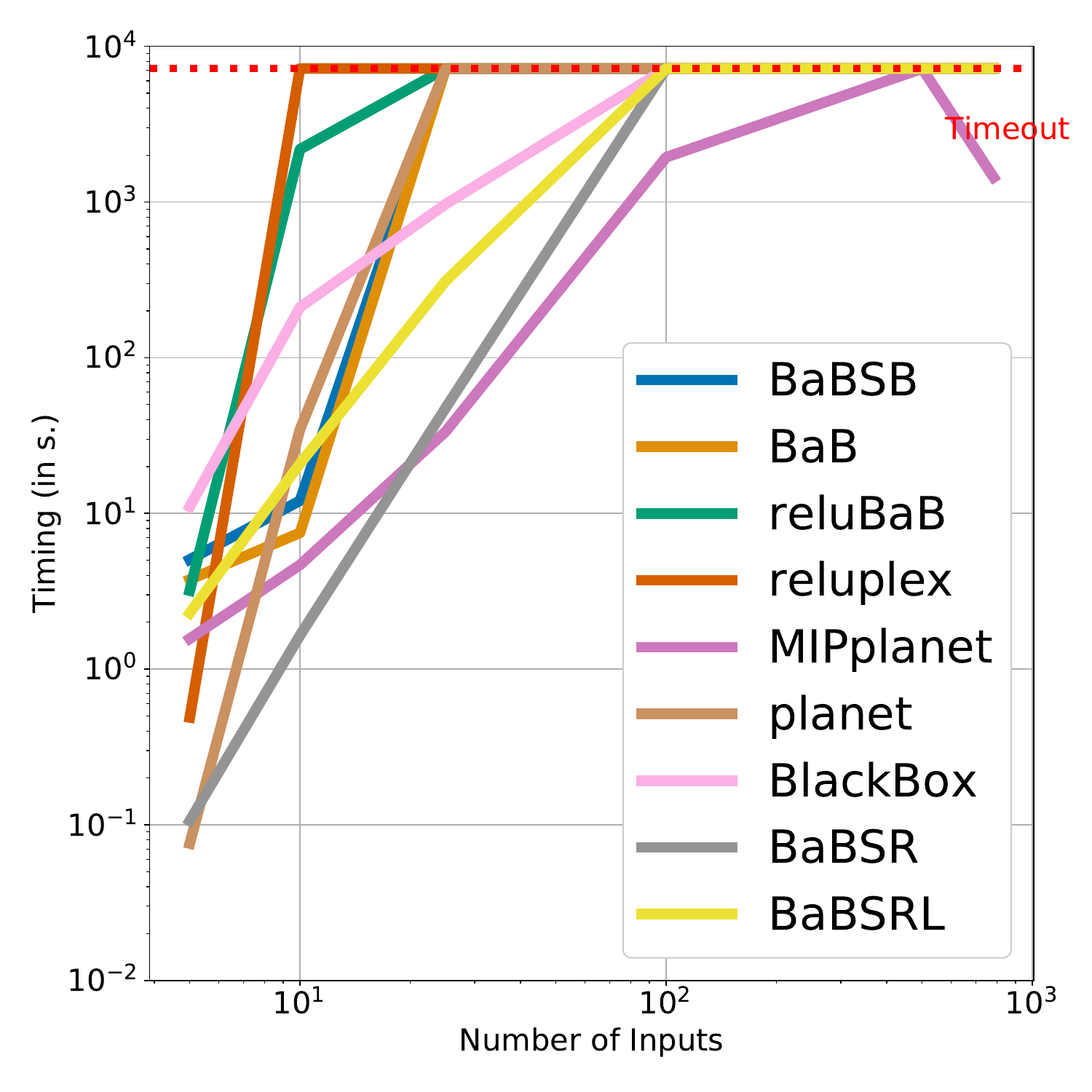}
    \caption{\label{fig:inp-ana}Number of inputs}
    \vspace{-2pt}
  \end{subfigure}%
  \hspace{.15\textwidth}%
  \begin{subfigure}{.35\textwidth}
    \includegraphics[width=\textwidth]{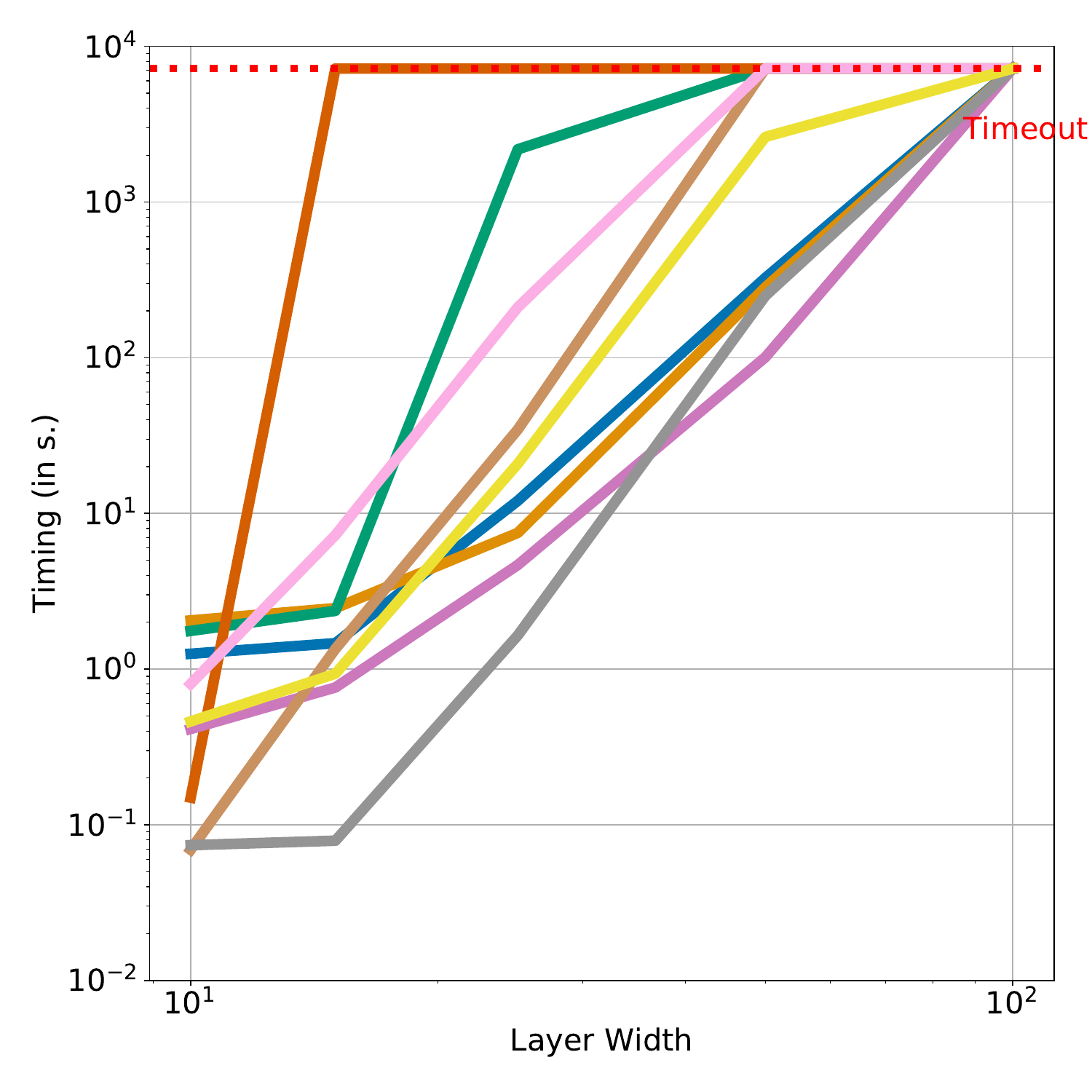}
    \caption{\label{fig:width-ana}Layer width}
    \vspace{-2pt}
  \end{subfigure}%

  \begin{subfigure}{.35\textwidth}
    \includegraphics[width=\textwidth]{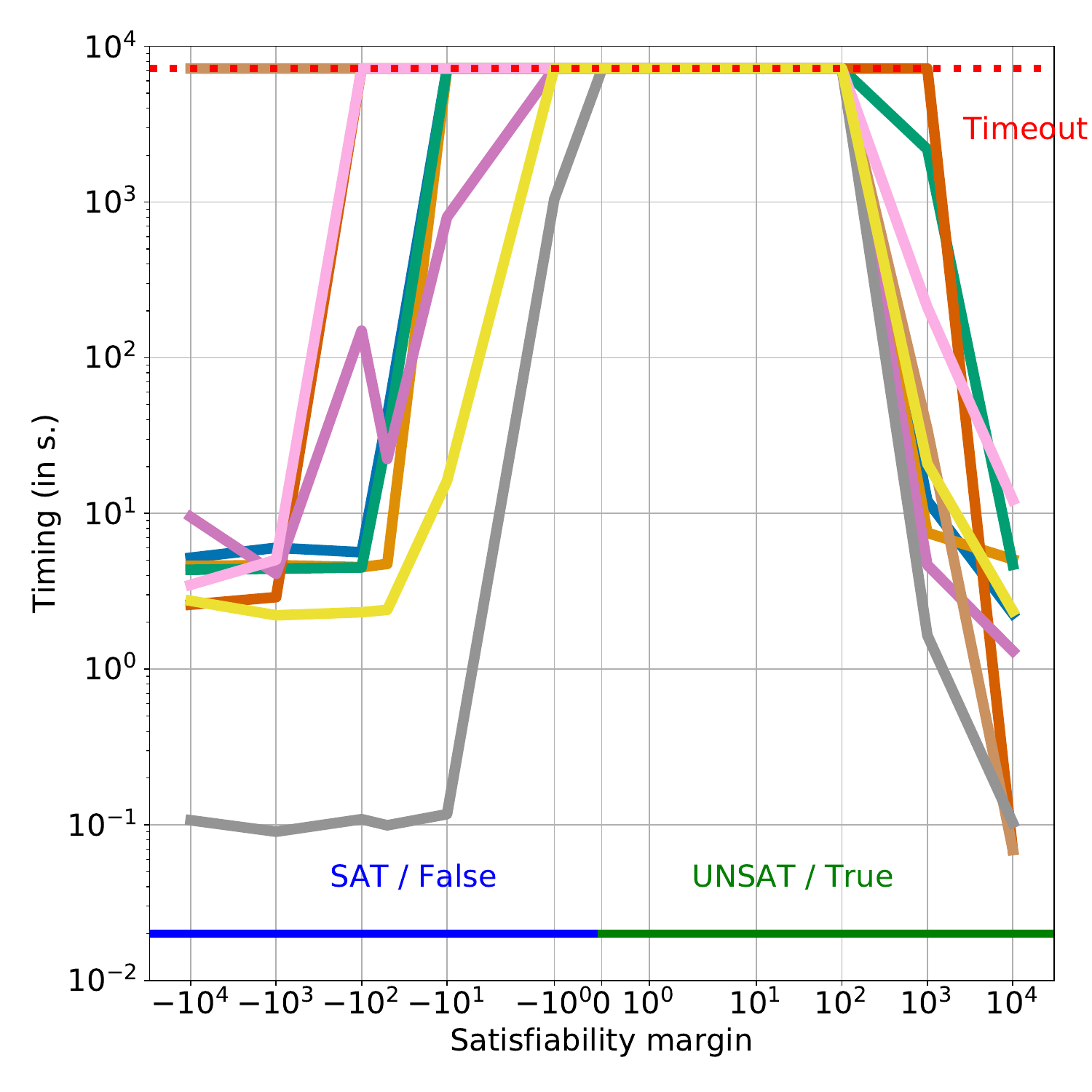}
    \caption{\label{fig:margin-ana}Margin}
    \vspace{-5pt}
  \end{subfigure}%
  \hspace{.15\textwidth}%
  \begin{subfigure}{.35\textwidth}
    \includegraphics[width=\textwidth]{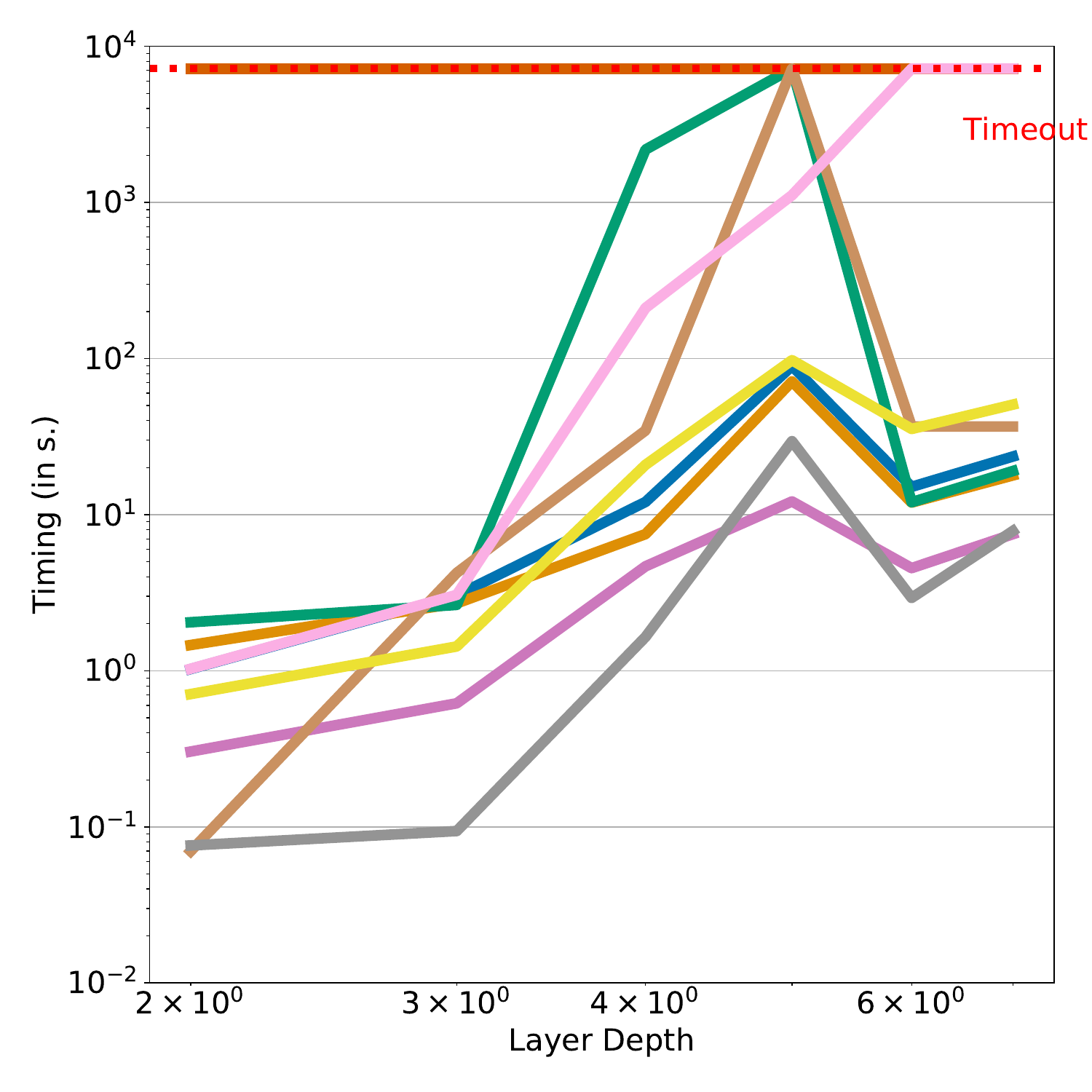}
    \caption{\label{fig:depth-ana}Network depth}
    \vspace{-5pt}
  \end{subfigure}%

  \caption{\label{fig:hyp-analysis} Impact of the various parameters over
    the runtimes of the different solvers. The base network has 10 inputs and 4
    layers of 25 hidden units, and the property to prove is true with a margin
    of 1000. Each of the plots corresponds to a variation of one of this
    parameters.}
    \vspace{-20pt}
\end{figure}

Finally, we study how the performance of each method is impacted by various parameters. Firstly, consider the case of various network architectures of the PCAMNIST data set. In the graphs of Figure~\ref{fig:hyp-analysis}, the trend for almost all methods
are similar, which seems to indicate that hard properties are intrinsically hard
and not just hard for a specific solver. Figure~\ref{fig:inp-ana} shows an
expected trend: the larger the number of inputs, the harder the problem is.
Similarly, Figure~\ref{fig:width-ana} shows unsurprisingly that wider networks
require more time to solve, which can be explained by the fact that they have
more non-linearities. The impact of the margin, as shown in
Figure~\ref{fig:margin-ana} is also clear. Properties that are true or false
with large satisfiability margin are easy to prove, while properties that have
small satisfiability margins are significantly harder. It is interesting to see the inconsistent performance of \textbf{MIPplanet}, which could be due to the different strategies used by Gurobi for different sized problems. In addition, consistent results of \textbf{BaBSR} outperforming \textbf{MIPplanet} on easy problems can be observed. We point out that the PCAMNIST data set is a small data set with only 27 networks. Observations should be made with care in terms of generalisability.  \\

\begin{figure}
 \vspace{-20pt}
  \begin{minipage}[c]{.5\textwidth}
    \includegraphics[width=.95\textwidth]{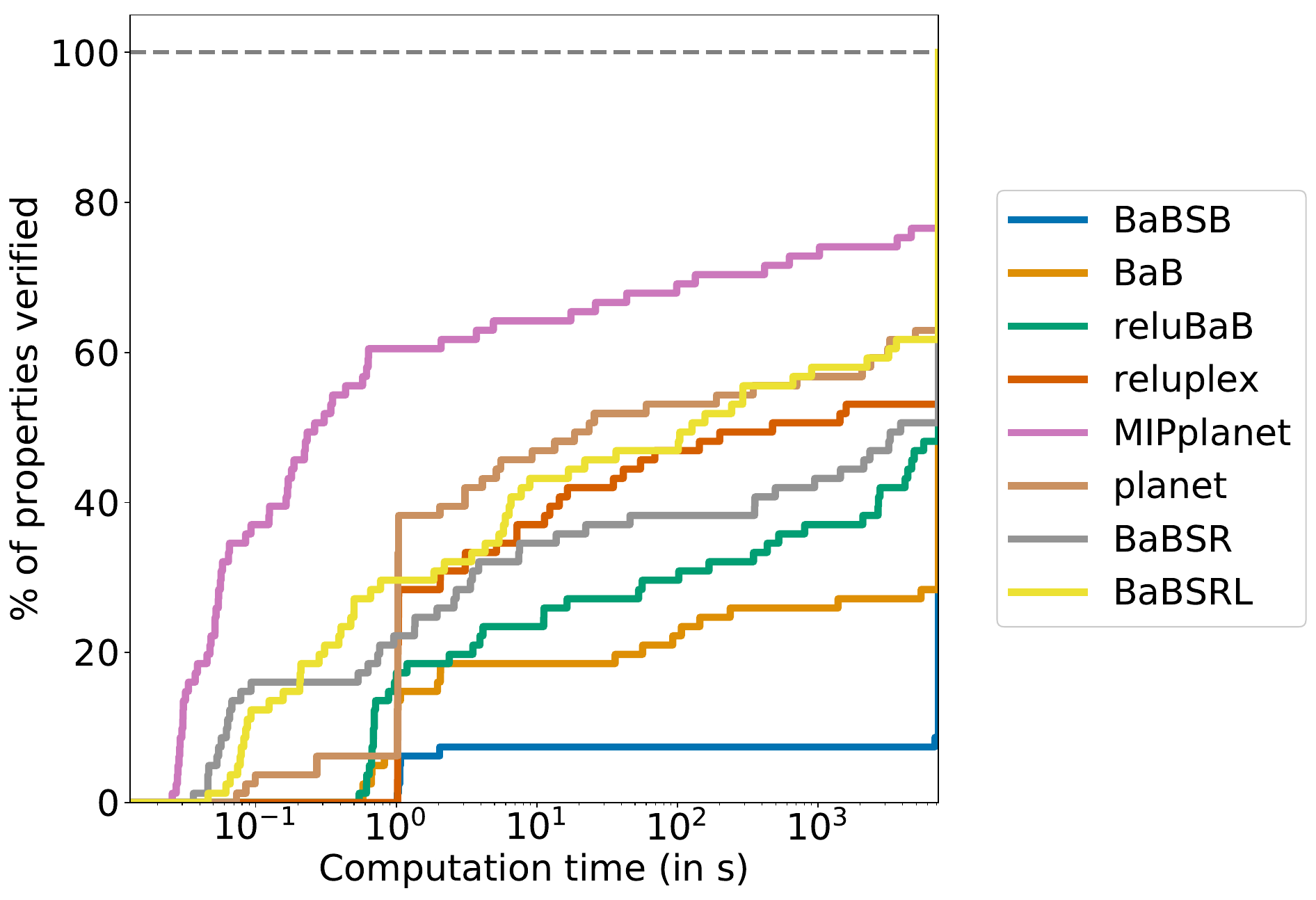}
  \end{minipage}
  \hfill
  \begin{minipage}[c]{.45\textwidth}
    \caption{\label{fig:twin_analysis} By design, the TwinStream data set consists of UNSAT properties only. All methods returned correct results for properties verified apart from \textbf{Reluplex}, which return SAT for several properties. These properties are treated as unsolved for these methods. \textbf{MIPplanet} outperforms all other methods on the TwinStream data set.}
  \end{minipage}
    \vspace{-10pt}
\end{figure}

The TwinStream data set introduces a possible standpoint to take when selecting splitting methods and bounding methods for BaB algortihms or other non-BaB methods to best solve the properties at hand. In Figure~\ref{fig:twin_analysis}, we see that \textbf{MIPplanet} performed the best over all properties and input split methods \textbf{BaBSB} and \textbf{BaB} performed the worst. Despite the fact that all twinladder networks are small, ReLU split strategy should be preferred when highly correlated layers are present. In terms of ReLU split based methods, the method with the tightest relaxation \textbf{reluBaB} and the method with the supposedly effective ReLU prioritizing heuristics \textbf{BaBSR} performed the worst. This is expected, as when the correlation among the hidden nodes of the same layer is high (by which we mean if we write each node as a linear combination of hidden nodes on the previous layer, the coefficient vectors are highly correlated), the ReLU score used in \textbf{BaBSR} will become way too loose to be of any use. For example, if several ReLU nodes $x_{i[p]}$ are highly correlated with the node $x_{i[j]}$, forcing $\xhat_{i[j]}\leq 0$ or $\xhat_{i[j]}\geq 0$ will lead to notable changes to the upper or lower bound of $x_{i[p]}$ respectively. In extreme cases, it means $\xhat_{i[p]}\leq 0$ or $\xhat_{i[p]}\geq 0$ if the weights associate with $x_{i[j]}$ and $x_{i[k]}$ have a correlation of one. Estimating improvements by keeping all other terms the same is thus unreasonable in this case. The \textbf{reluBaB} method mainly suffers from its computational cost. The method \textbf{BaBSRL} lies between \textbf{BaBSR} and \textbf{reluBaB}. Since only one layer is updated via the LP, \textbf{BaBSRL} has tighter relaxations than \textbf{BaBSR} so the ReLU prioritising heuristic of \textbf{BaBSR} can make better branching decisions in the following steps. Yet, the computational cost of \textbf{BaBSRL} is much lower than that of \textbf{reluBaB}. An improved performance of \textbf{BaBSRL} over \textbf{BaBSR} and \textbf{reluBaB} can be observed. Overall, on the small networks with highly correlated layers like networks of Twinstream, \textbf{MIPplanet} is a definite winner. However, as we have observed previously, \textbf{MIPplanet} is likely to suffer from scalability issues. We expect \textbf{BaBSRL} might be a better option on large networks with highly correlated layers.

\section{Conclusion}
Through the lens of a unified Branch-and-Bound framework, we have identified the weakness of existing methods and proposed new methods to correct and improve on it. These new methods are effective by achieving considerate performance enhancements on our comprehensive data sets. However, there is still much room for improvements. We illustrate a few based on the BaB framework. In terms of bounding strategies, we observe tighter intermediate bounds could lead to faster convergence but they are expensive to obtain. Given the layer-wise structure of neural networks, exploiting the GPU computing power might be a potential way to compute tighter intermediate bounds cheaply. For branching strategies, methods discussed mainly rely on heuristics, which are likely to fail when problem changes. Recent studies have shown learning heuristics through graph neural networks might overcome the issue but they have high offline cost. Cheaper while effective strategies should be possible. Finally, BaB based methods require solving LPs, the main bottleneck hindering their development. Since LPs are mostly solved to decide whether a branching should be conducted on a (sub-)problem, learning to imitate LP decision could be a potential way to speed up the whole process by orders of magnitudes. 
\par
We encourage the development of new methods to address these issues and hope our inclusive and various data sets could facilitate the process by allowing comprehensive evaluations and comparison studies.


\acks{This work was supported by ERC grant ERC-2012-AdG 321162-HELIOS, EPSRC grant Seebibyte EP/M013774/1 and EPSRC/MURI grant EP/N019474/1. We would also like to acknowledge the Royal Academy of Engineering and FiveAI.}

\appendix
\section*{Appendix A. Planet Approximation}
The feasible set of the Mixed Integer Programming formulation is given by the
following set of equations. We assume that all $\mbf{l_i}$ are negative and
$\mbf{u_i}$ are positive. In case this isn't true, it is possible to just update
the bounds such that they are.
\begin{subequations}
\vskip -20pt
  \begin{align}
    &\mbf{l_0} \leq \mbf{x_0} \leq \mbf{u_0}& \label{eq:ctx-bounds1}\\
    &\mbf{\xhat_{i+1}} = W_{i+i} \mbf{x_i} + \mbf{b_{i+i}} &\quad \forall i \in \{0, \ n-1\}\label{eq:lin-op1}\\
    &\mbf{x_i} \geq 0 &\quad \forall i \in \{1, \ n-1\}\\
    &\mbf{x_i} \geq \mbf{\xhat_i}&\quad \forall i \in \{1, \ n-1\}\\
    &\mbf{x_i} \leq \mbf{\xhat_i} - \mbf{l_i}\cdot(1 - \bm{\delta_i})&\quad \forall i \in \{1, \ n-1\}\\
    &\mbf{x_i} \leq \mbf{u_i} \cdot \bm{\delta_i}&\quad \forall i \in \{1, \ n-1\}\\
    &\bm{\delta_i} \in \{0, 1\}^{h_i}&\quad \forall i \in \{1, \ n-1\}\\
    &\xhat_{n} \leq 0. \label{eq:final1}
  \end{align}
  \label{eq:mipsat-problem1}
\end{subequations}

The level 0 of the Sherali-Adams hierarchy of relaxation \citet{Sherali1994}
doesn't include any additional constraints. Indeed, polynomials of degree 0 are
simply constants and their multiplication with existing constraints followed by
linearization therefore does not add any new constraints. As a result, the
feasible domain given by the level 0 of the relaxation corresponds simply to the
removal of the integrality constraints:
\begin{subequations}
  \begin{align}
    &\mbf{l_0} \leq \mbf{x_0} \leq \mbf{u_0}& \label{eq:ctx-bounds2}\\
    &\mbf{\xhat_{i+1}} = W_{i+i} \mbf{x_i} + \mbf{b_{i+i}} &\quad \forall i \in \{0, \ n-1\}\label{eq:lin-op2}\\
    &\mbf{x_i} \geq 0 &\quad \forall i \in \{1, \ n-1\}\\
    &\mbf{x_i} \geq \mbf{\xhat_i}&\quad \forall i \in \{1, \ n-1\}\\
    &\mbf{x_i} \leq \mbf{\xhat_i} - \mbf{l_i}\cdot(1 - \bm{d_i})&\quad \forall i \in \{1, \ n-1\}\label{eq:xub1}\\
    &\mbf{x_i} \leq \mbf{u_i} \cdot \bm{d_i}&\quad \forall i \in \{1, \ n-1\}\label{eq:xub2}\\
    & \underline{0 \leq \bm{d_i} \leq 1 }&\quad \forall i \in \{1, \ n-1\}\\
    &\xhat_{n} \leq 0. \label{eq:final2}
  \end{align}
  \label{eq:mipsat-problem}
\end{subequations}

Combining the Equations~\ref{eq:xub1} and \ref{eq:xub2}, looking at a single
unit $j$ in layer $i$, we obtain:
\begin{equation}
  x_{i[j]} \leq \min\left( \xhat_{i[j]} - l_i (1 - d_{i[j]}), u_{i[j]} d_{i[j]} \right).
  \label{eq:x-minub}
\end{equation}
The function mapping $d_{i[j]}$ to an upperbound of $x_{i[j]}$ is a minimum of
linear functions, which means that it is a concave function. As one of them is
increasing and the other is decreasing, the maximum will be reached when they
are both equals.

\begin{equation}
  \begin{split}
    \xhat_{i[j]} - l_{i[j]} (1 - d^{\star}_{i[j]}) &= u_{i[j]}d^{\star}_{i[j]}\\
    \Leftrightarrow \qquad d^{\star}_{i[j]} &= \frac{\xhat_{i[j]} - l_{i[j]}}{u_{i[j]}-l_{i[j]}}.
  \end{split}
\end{equation}

Plugging this equation for $d^{\star}$ into Equation~\ref{eq:x-minub} gives
that:
\begin{equation}
    x_{i[j]} \leq u_{i[j]} \frac{\xhat_{i[j]} - l_{i[j]}}{u_{i[j]} - l_{i[j]}},
\end{equation}
which corresponds to the upper bound of $x_{i[j]}$ introduced for Planet~\citep{Ehlers2017}.

\begin{figure}
  \centering
  \begin{tikzpicture}
  \tikzset{dummy/.style= {inner sep=0, outer sep=0}}
  \tikzset{cross/.style={cross out, draw,
      minimum size=3*(#1-\pgflinewidth),
      inner sep=0pt, outer sep=0pt,
      thick}}

  \draw[-, ultra thick](-1, 0) to (0, 0) to (1, 1);

  \draw[dashed](-1, -0.5) to (-1, 1.5);
  \draw[dashed](1, -0.5) to (1, 1.5);

  \draw[fill=green](-1, 0) -- (0,0) -- (1, 1);

  \node[cross=2pt] at (-1, 0) {};
  \node[dummy](lb-lab) at (-1.3, -0.3) {$l_{i[j]}$};
  \node[cross=2pt] at (1, 0) {};
  \node[dummy](ub-lab) at (1.35, -0.3) {$u_{i[j]}$};

  \draw[-latex](-1.5,0) to (2, 0);
  \node[dummy](x-label) at (2.3, 0) {$\xhat_{i[j]}$};
  \draw[-latex](0,-0.5) to (0, 1.5);
  \node[dummy](x-label) at (0, 1.8) {$x_{i[j]}$};

\end{tikzpicture}
  \caption{\label{fig:relu-hull} Feasible domain corresponding to the
    Planet relaxation for a single ReLU.}
\end{figure}
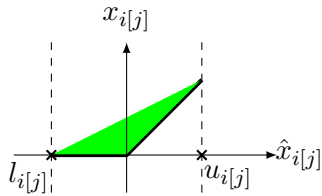

\section*{Appendix B. MaxPooling}
For space reason, we only described the case of ReLU activation function in the
main paper. We now present how to handle MaxPooling activation, either by
converting them to the already handled case of ReLU activations or by
introducing an explicit encoding of them when appropriate.

\subsection*{B.1 Mixed Integer Programming}
Similarly to the encoding of ReLU constraints using binary variables and bounds
on the inputs, it is possible to similarly encode MaxPooling constraints.
The constraint
\begin{equation}
  y = \max\left( x_1, \dots, x_k \right)
\end{equation}
can be replaced by
\begin{subequations}
  \begin{align}
    &y \geq x_i \qquad &&\forall i \in \{1\dots k\}\\
    &y \leq x_i + (u_{x_{1:k}} - l_{x_i})(1 - \delta_i) \quad &&\forall i \in \{1\dots k\}\\
    &\sum_{i \in \{1\dots k\}}  \delta_i = 1 \\
    &\delta_i \in \{0, 1\} \quad &&\forall i \in \{1\dots k\}.
  \end{align}
\end{subequations}
where $u_{x_{1:k}}$ is an upper-bound on all $x_{i}$ for $i \in \{1 \dots k\}$ and $l_{x_i}$ is a lower bound on $x_{i}$.

\subsection*{B.2 Reluplex}
In the version introduced by \citep{Katz2017}, there is no support for
MaxPooling units. As the canonical representation we evaluate needs them, we
provide a way of encoding a MaxPooling unit as a combination of Linear function
and ReLUs.
\par
To do so, we decompose the element-wise maximum into a series of pairwise
maximum
\begin{equation}
  \begin{split}
    \max\left( x_j, x_2, x_3, x_4 \right) = \max(\ &\max\left( x_1, x_2 \right),\\
      & \max\left( x_3, x_4 \right) )
  \end{split}
\end{equation}
and decompose the pairwise maximums as sum of ReLUs:
\begin{equation}
  \max\left( x_1, x_2 \right) = \max\left( x_1 - x_2, \ 0 \right) + \max\left( x_2 - l_{x_2}, 0 \right) + l_{x_2},
\end{equation}
where $l_{x_2}$ is a pre-computed lower bound of the value that $x_2$
can take.
\par
As a result, we have seen that an elementwise maximum such as a MaxPooling unit
can be decomposed as a series of pairwise maximum, which can themselves be
decomposed into a sum of ReLUs units. The only requirement is to be able to
compute a lower bound on the input to the ReLU, for which the methods discussed
in the paper can help.

\subsection*{B.3 Planet}
As opposed to Reluplex, Planet~\citet{Ehlers2017} directly supports MaxPooling units.
The SMT solver driving the search can split either on ReLUs, by considering
separately the case of the ReLU being passing or blocking. It also has the
possibility on splitting on MaxPooling units, by treating separately each
possible choice of units being the largest one.
\par 
For the computation of lower bounds, the constraint
\begin{equation}
  y = \max\left( x_1, x_2, x_3, x_4 \right)
\end{equation}
is replaced by the set of constraints:
\begin{subequations}
  \begin{align}
    y &\geq \text{x}_i \qquad \forall i \in \{1 \dots 4\}\\
    y &\leq \sum_i \left( x_i - l_{x_i} \right) + \max_i l_{x_i},
  \label{eq:maxpool-cvxhull}
  \end{align}
\end{subequations}
where $x_i$ are the inputs to the MaxPooling unit and $l_{x_i}$ their lower
bounds.

\section*{Appendix C. Mixed Integers Variants}

\subsection*{C.1 Encoding}
Several variants of encoding are available to use Mixed Integer Programming as a
solver for Neural Network Verification. As a reminder, in the main paper we used
the formulation of \citet{Tjeng2019}:
\begin{subequations}
  \begin{flalign}
    x_i = \max\left(\mbf{\xhat_i}, 0\right) \quad \Rightarrow \quad \bm{\delta_i} \in
    \{0,1\}^{h_i}, &\quad\mbf{x_i} \geq 0, \ \qquad  \mbf{x_i} \leq \mbf{u_i} \cdot \bm{\delta_i}\label{eq:mip_0a}\\
    & \quad \mbf{x_i}\geq \mbf{\xhat_i},   \qquad \mbf{x_i} \leq \mbf{\xhat_i} - \mbf{l_i}\cdot(1 - \bm{\delta_i}).\label{eq:mip_inpa}
  \end{flalign}
  \label{eq:mip-form}%
\end{subequations}

An alternative formulation is the one of \citet{Lomuscio2017} and \citet{Cheng2017}:
\begin{subequations}
  \begin{flalign}
    x_i = \max\left(\mbf{\xhat_i}, 0\right) \quad \Rightarrow \quad \bm{\delta_i} \in
    \{0,1\}^{h_i}, &\quad\mbf{x_i} \geq 0, \ \qquad  \mbf{x_i} \leq \mbf{M_i} \cdot \bm{\delta_i}\label{eq:mip_0a1}\\
    & \quad \mbf{x_i}\geq \mbf{\xhat_i},   \qquad \mbf{x_i} \leq \mbf{\xhat_i} - \mbf{M_i}\cdot(1 - \bm{\delta_i}).\label{eq:mip_inpa1}
  \end{flalign}
  \label{eq:altmip-form}%
\end{subequations}
where $\mbf{M_i} = \max\left( -\mbf{l_i}, \mbf{u_i} \right)$. This is
fundamentally the same encoding but with a sligthly worse bounds that is used,
as one of the side of the bounds isn't as tight as it could be.

\subsection*{C.2 Obtaining Bounds}
The formulation described in Equations~\ref{eq:mip-form} and \ref{eq:altmip-form} are dependant on obtaining lower and upper bounds for
the value of the activation of the network.

\subsubsection*{C.2.1 Interval Analysis}
One way to obtain them, mentionned in the paper, is the use of interval
arithmetic~\citep{Hickey2001}. If we have bounds $\mbf{l_{i}}, \mbf{u_{i}}$ for a vector
$\mbf{x_{i}}$, we can derive the bounds $\mbf{\hat{l}_{i+1}}, \mbf{\hat{u}_{i+1}}$ for a vector
\mbox{$\mbf{\xhat_{i+1}} = W_{i+1} \mbf{x_i} + b_{i+1}$}

\begin{subequations}
  \begin{align}
    \hat{l}_{i+1[j]} &= \sum_k \left( W_{i+1[j, k]}^+ l^+_{i[k]} + W_{i+1[j,k]}^- u^+_{i[k]} \right) + b_{i+1[j]}\\
    \hat{u}_{i+1[j]} &= \sum_k \left( W_{i+1[j, k]}^+ u^+_{i[k]} + W_{i+1[j,k]}^- l^+_{i[k]} \right) + b_{i+1[j]}
  \end{align}
  \label{eq:interval-anal}%
\end{subequations}
with the notation \mbox{$a^+=\max(a, 0)$} and \mbox{$a^-=\min(a, 0)$}.
Propagating the bounds through a ReLU activation is simply equivalent to
applying the ReLU to the bounds.

\subsubsection*{C.2.2 Planet Linear approximation}
An alternative way to obtain bounds is to use the relaxation of Planet. This is
the methods that was employed in the paper: we build incrementally the network
approximation, layer by layer. To obtain the bounds over an activation, we
optimize its value subject to the constraints of the relaxation.

Given that this is a convex problem, we will achieve the optimum. Given that it
is a relaxation, the optimum will be a valid bound for the activation (given
that the feasible domain of the relaxation includes the feasible domains subject
to the original constraints).

Once this value is obtained, we can use it to build the relaxation for the
following layers. We can build the linear approximation for the whole network
and extract the bounds for each activation to use in the encoding of the MIP.
While obtaining the bounds in this manner is more expensive than simply doing
interval analysis, the obtained bounds are better.

\subsection*{C.2 Objective Function}
In the paper, we have formalised the verification problem as a satisfiability
problem, equating the existence of a counterexample with the feasibility of the
output of a (potentially modified) network being negative.
\par 
In practice, it is beneficial to not simply formulate it as a feasibility
problem but as an optimization problem where the output of the network is
explicitly minimized.

\subsection*{C.3 Comparison}
We present here a comparison on CollisionDetection and ACAS of the different
variants.
\begin{enumerate}
  \item \textbf{Planet-feasible} uses the encoding of
    Equation~\ref{eq:mip-form}, with bounds obtained based on the planet
    relaxation, and solve the problem simply as a satisfiability problem.
  \item \textbf{Interval} is the same as \textbf{Planet-feasible}, except that
    the bounds used are obtained by interval analysis rather than with the
    Planet relaxation.
  \item \textbf{Planet-symfeasible} is the same as \textbf{Planet-feasible},
    except that the encoding is the one of Equation~\ref{eq:altmip-form}.
  \item \textbf{Planet-opt} is the same as \textbf{Planet-feasible}, except that
    the problem is solved as an optimization problem. The MIP solver attempt to
    find the global minimum of the output of the network. Using Gurobi's
    callback, if a feasible solution is found with a negative value, the
    optimization is interrupted and the current solution is returned. This
    corresponds to the version that is reported in the main paper.
  \end{enumerate}
The comparison also includes two variants of \textbf{BlackBox}: \textbf{BlackBox}
and \textbf{BlackBoxNoOpt}. Similarly to \textbf{Planet-opt}, \textbf{BlackBox}
attempts to do global optimization and interrupt the search when a feasible
solution with negative value is found. \textbf{BlackBoxNoOpt} works like the
other MIP encoding by simply encoding the problem as satisfiability.

\begin{figure}
  \hfill%
  \begin{subfigure}[t]{.4 \textwidth}
    \includegraphics[width=\textwidth]{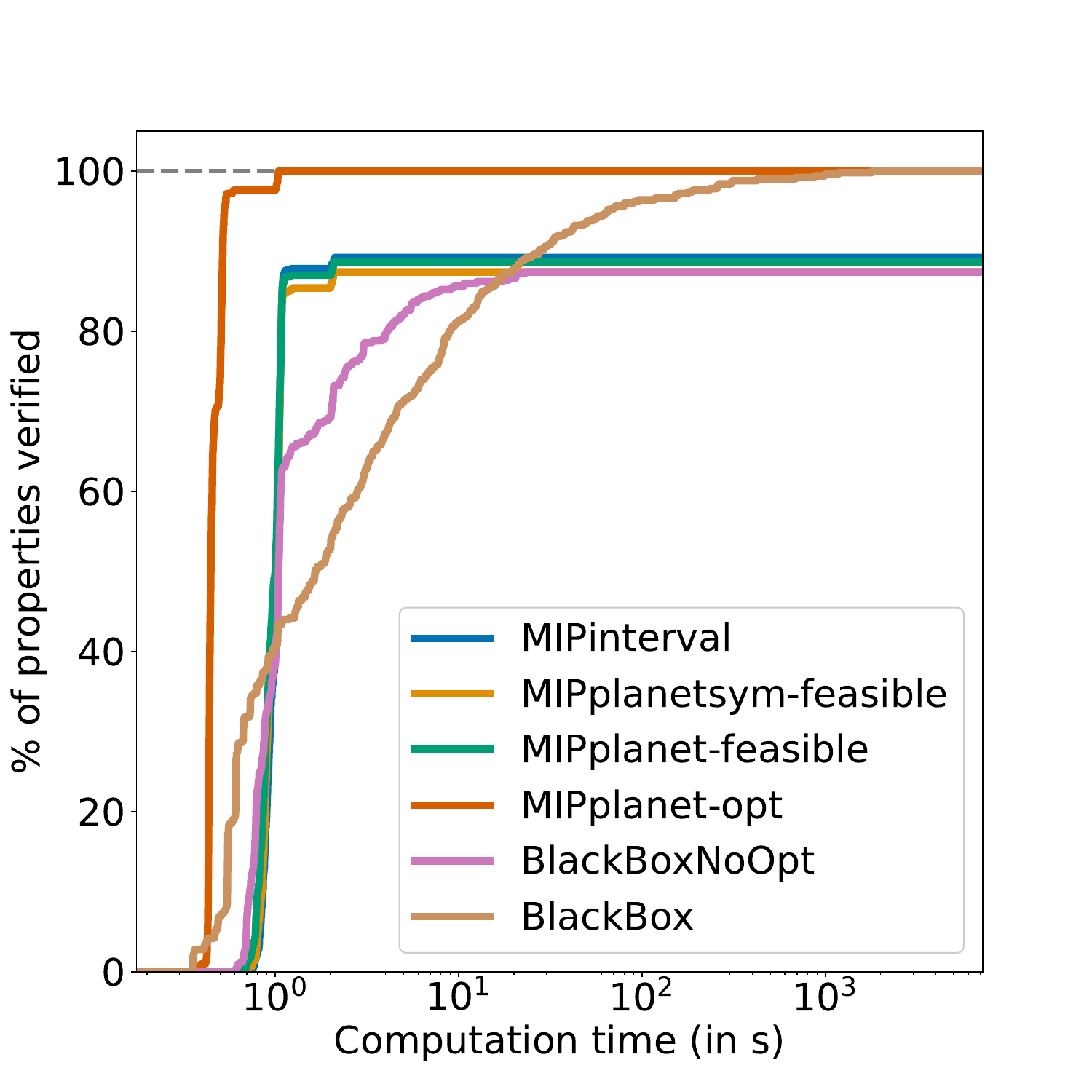}
    \caption{\label{fig:mip-collision-comparison} CollisionDetection data set}
  \end{subfigure}%
  \hfill%
  \begin{subfigure}[t]{.4 \textwidth}
    \includegraphics[width=\textwidth]{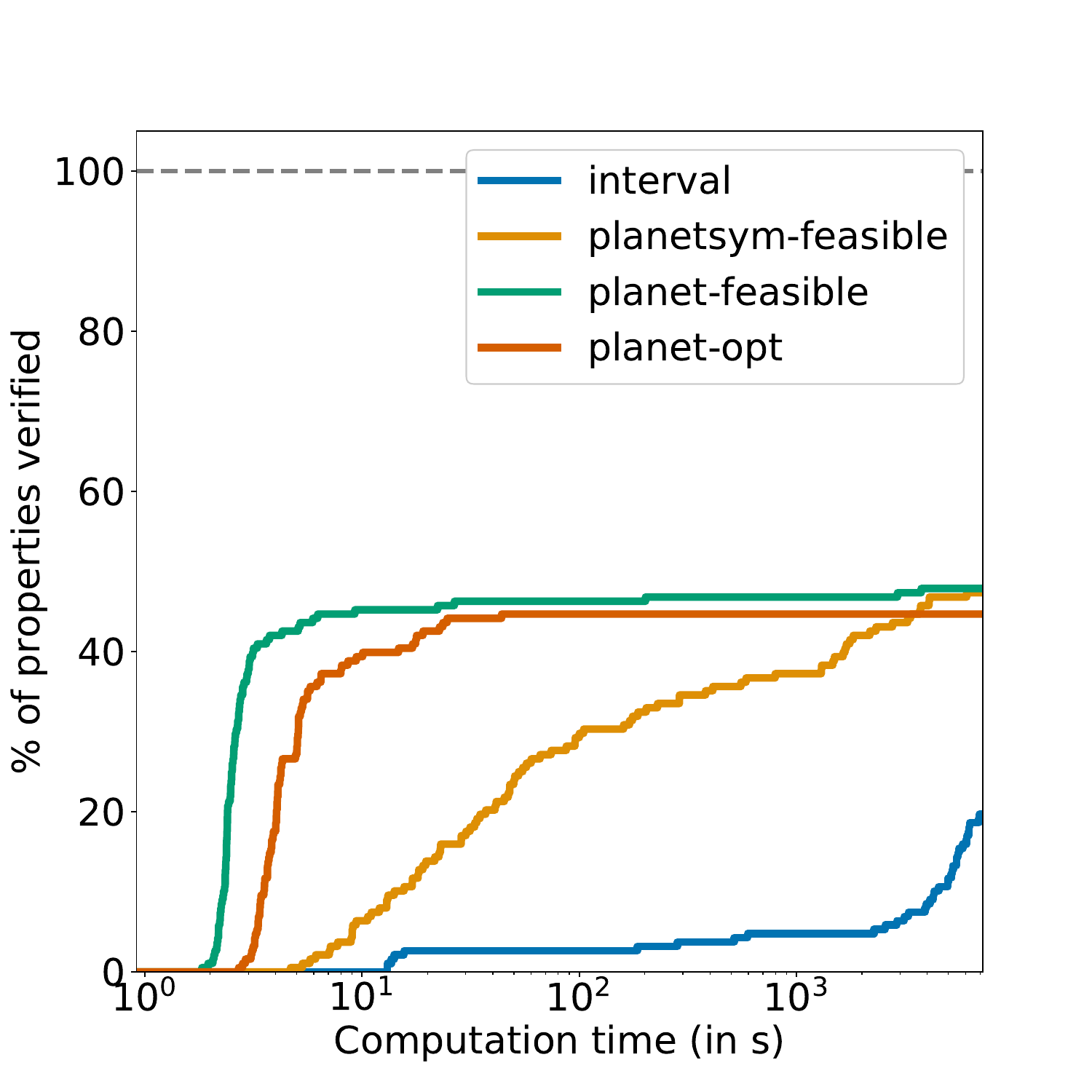}
    \caption{\label{fig:mip-acas-comparison} ACAS data set}
  \end{subfigure}%
  \hfill
  \caption{\label{fig:mip-comparison} Comparison between the different
      variants of MIP formulation for Neural Network verification.}
\end{figure}
The first observation that can be made is that when we look at the
CollisionDetection data set in Figure~\ref{fig:mip-collision-comparison}, only
\textbf{Planet-opt} and \textbf{BlackBox} solves the data set to 100\% accuracy.
The reason why the other methods don't reach it is not because of timeout but
because they return spurious counterexamples. As they encode only satisfiability
problem, they terminate as soon as they identify a solution with a value of
zero. Due to the large constants involved in the big-M, those solutions are
sometimes not actually valid counterexamples. This is a significant advantage to
encoding the problem as optimization problems versus simply as satisfiability problems.
\par
The other results that we can observe is the impact of the quality of the bounds
when the networks get deeper, and the problem becomes therefore more complex,
such as in the ACAS data set. \textbf{Interval} has the worst bounds and is much
slower than the other methods. \textbf{Planetsym-feasible}, with its slightly
worse bounds, performs worse than \textbf{Planet-feasible} and \textbf{Planet-opt}.

\section*{Appendix D. PCAMNIST Details}
\textbf{PCAMNIST} is a novel data set that we introduce to get a better
understanding of what factors influence the performance of various methods. It
is generated in a way to give control over different architecture parameters.
The networks takes $k$ features as input, corresponding to the first $k$
eigenvectors of a Principal Component Analysis decomposition of the digits from
the MNIST data set. We also vary the depth (number of layers), width (number of
hidden unit in each layer) of the networks. We train a different network for
each combination of parameters on the task of predicting the parity of the
presented digit. This results in the accuracies reported in
Table~\ref{tab:PCAMNIST-acc}.
\par
The properties that we attempt to verify are whether there exists an input for
which the score assigned to the odd class is greater than the score of the even
class plus a large confidence. By tweaking the value of the confidence in the
properties, we can make the property either true or false, and we can choose by
how much is it true. This gives us the possibility of tweaking the ``margin'',
which represent a good measure of difficulty of a network.
\par
\begin{figure}
  \centering
  \includegraphics[width=.5\textwidth]{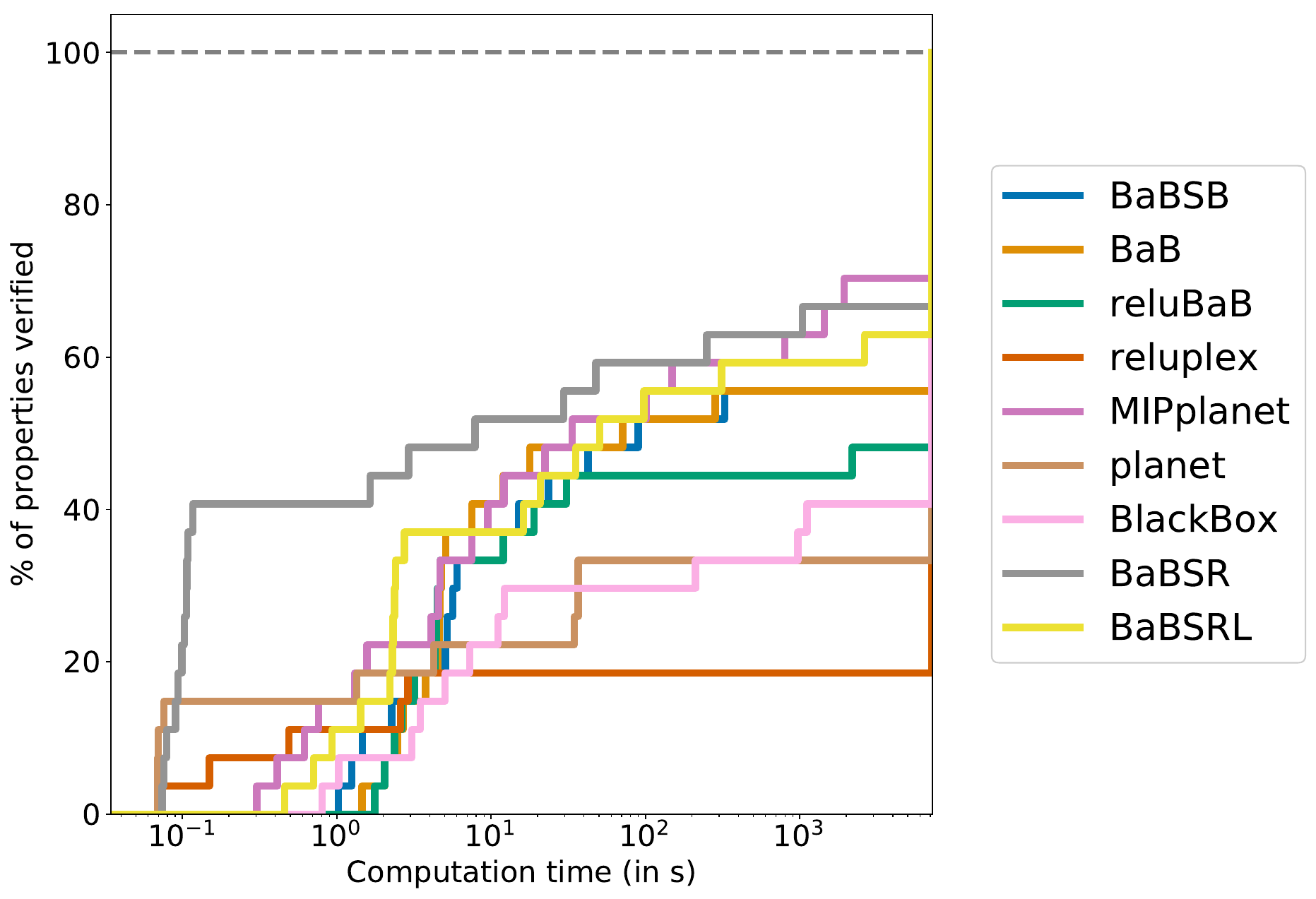}
  \caption{\label{fig:cactus_pcamnist}Proportion of properties verifiable
      for varying time budgets depending on the methods employed. Overall, \textbf{BaBSR} performed the best on easy properties but worse than \textbf{MIPplanet} on difficult properties, which is consistent to what is observed on properties of the reduced Robust network.}
\end{figure}

 
In addition to the impact of each factors separately as was shown in the main
paper, we can also look at it as a generic data set and plot the cactus plots
like for the other data sets. This can be found in Figure~\ref{fig:cactus_pcamnist}.


\label{app:pcamnist-acc}
\begin{table}[h]
  \center
  \resizebox{.48\textwidth}{!}{
    \begin{tabular}{ccc@{\hspace*{4ex}}|cc}
      \toprule
      \multicolumn{3}{c|}{\textbf{Network Parameter}} & \multicolumn{2}{c}{\textbf{Accuracy}}\\
      \textbf{Nb inputs} & \textbf{Width} & \textbf{Depth} & \textbf{Train} & \textbf{Test} \\
      \midrule
      5 & 25 & 4 & 88.18\% & 87.3\%\\
      10 & 25 & 4 & 97.42\% & 96.09\%\\
      25 & 25 & 4 & 99.87\% & 98.69\%\\
      100 & 25 & 4 & 100\% & 98.77\%\\
      500 & 25 & 4 & 100\% & 98.84\%\\
      784 & 25 & 4 & 100\% & 98.64\%\\
      \midrule
      10 & 10 & 4 & 96.34\% & 95.75\%\\
      10 & 15 & 4 & 96.31\% & 95.81\%\\
      10 & 25 & 4 & 97.42\% & 96.09\%\\
      10 & 50 & 4 & 97.35\% & 96.0\%\\
      10 & 100 & 4 & 97.72\% & 95.75\%\\
      \midrule
      10 & 25 & 2 & 96.45\% & 95.71\%\\
      10 & 25 & 3 & 96.98\% & 96.05\%\\
      10 & 25 & 4 & 97.42\% & 96.09\%\\
      10 & 25 & 5 & 96.78\% & 95.9\%\\
      10 & 25 & 6 & 95.48\% & 95.2\%\\
      10 & 25 & 7 & 96.81\% & 96.07\%\\
      \bottomrule
    \end{tabular}}
  \caption{\label{tab:PCAMNIST-acc}Accuracies of the network trained for the
    \mbox{PCAMNIST} data set.}
\end{table}

\section*{Appendix E. TwinStream Details}
The networks contain two separate streams, where each of the streams has the same architecture, weights, and inputs. The final layer of the network computes the difference between the
outputs of the two streams, and adds a positive bias term, which we will refer to as the margin, denoted as $m$. As a result, the output is always equal to the value of the final bias. We give explicit formulations of weights and biases of the TwinStream networks. Given a N-layer stream with weights $W^s_1, \dots, W^s_N$ and biases $b^s_1, \dots, b^s_{N-1}$, the corresponding N-layer TwinStream network consists of following weights $W_1, \dots, W_N$ and biases $b_1, \dots, b_N$:
\[
W_1 = \begin{bmatrix} W^s_1\\ W^s_1 \end{bmatrix}, \qquad W_i =\begin{bmatrix} W^s_i & \mbf{0} \\ \mbf{0} & W^s_i \end{bmatrix} \quad\text{for }i\in \{2,\dots, N-1\}, \qquad W_N = \begin{bmatrix} W^s_N\\-W^s_N \end{bmatrix}.
\]
\[
b_i = \begin{bmatrix} b^s_i \\ b^s_i \end{bmatrix} \quad \text{for }i\in \{2,\dots, N-1\}, \qquad b_N = m.
\]
\par
On each of those networks, we attempt to prove the true property that the output of the network is positive. We generate streams with random weights using Glorot initialisation \citet{Glorot2010}. Various TwinStream Networks are constructed by varying the depth, number of hidden units in each of the stream, number of inputs, and the value of the margin. Note that
as opposed to the other two data sets, the weights aren’t the result of an optimization process and
therefore may not be representative of real use-cases.

\section*{Appendix F. Additional Performance Details}
Given that there is a difference in the way verification works for
SAT problems vs. UNSAT problems, we report also comparison results on the subset
of data sorted by decision type.

\begin{figure}
  \begin{minipage}[b]{\textwidth}
    \vskip -10pt
    \centering
    \begin{subfigure}[t]{.50\textwidth}
      \vskip 0pt
      \includegraphics[width=.90\textwidth]{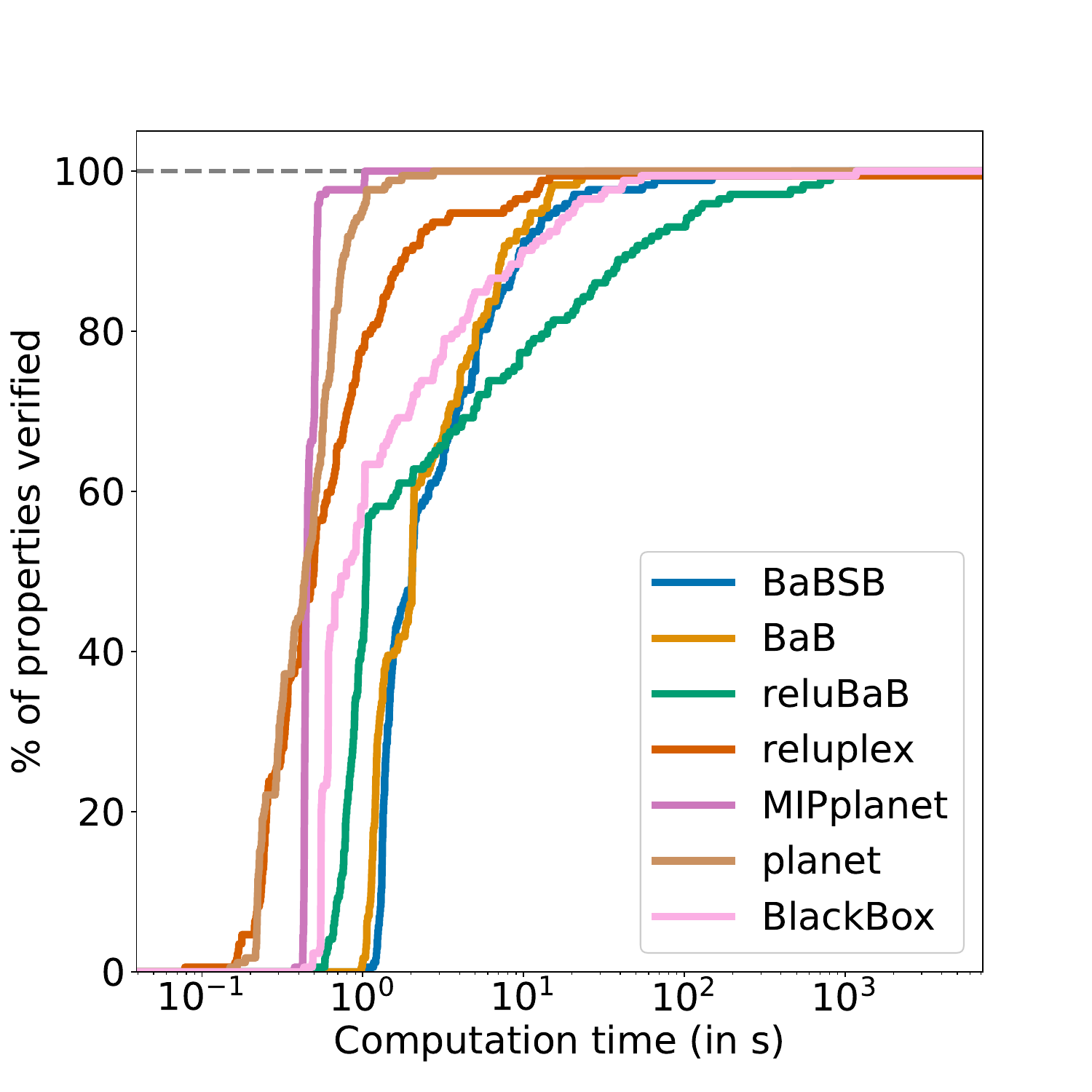}
      \caption{On SAT properties}
    \end{subfigure}%
    \begin{subfigure}[t]{.50\textwidth}
      \vskip 0pt
      \includegraphics[width=.90\textwidth]{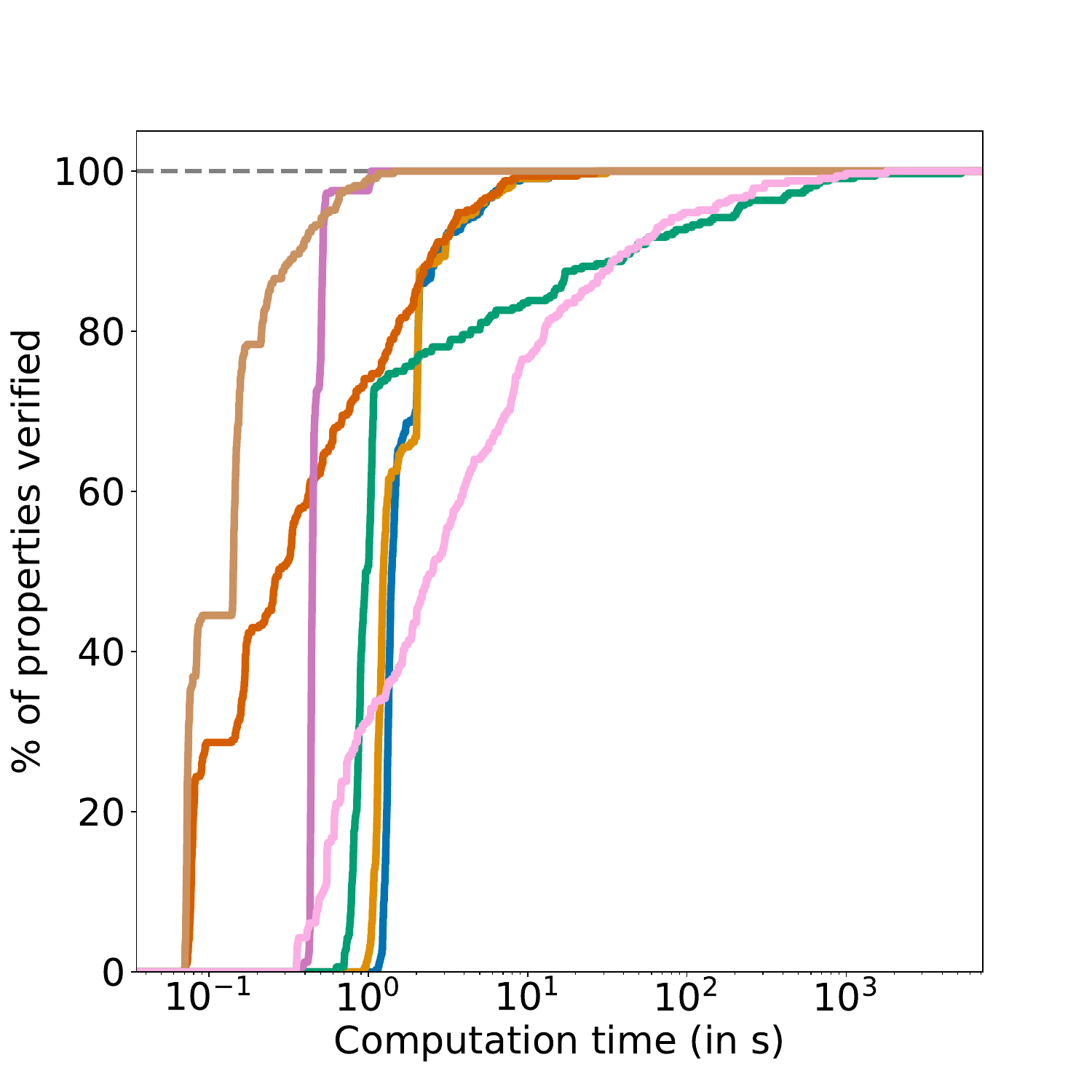}
      \caption{On UNSAT properties}
    \end{subfigure}
  \end{minipage}

  \begin{minipage}[t]{\textwidth}
    \vskip 0pt
    \caption{\label{fig:sat_unsat_collision}Proportion of properties
        verifiable by different methods under varying time budgets on the \textbf{CollisionDetection} data set. We can identify that all the errors that
        \textbf{BlackBox} makes are on SAT properties, as it returns incorrect counterexamples.}
  \end{minipage}
  \vskip -20pt
\end{figure}

\begin{figure}
  \begin{minipage}[b]{\textwidth}
    \centering
    \begin{subfigure}[t]{.50\textwidth}
      \vskip 0pt
      \includegraphics[width=.90\textwidth]{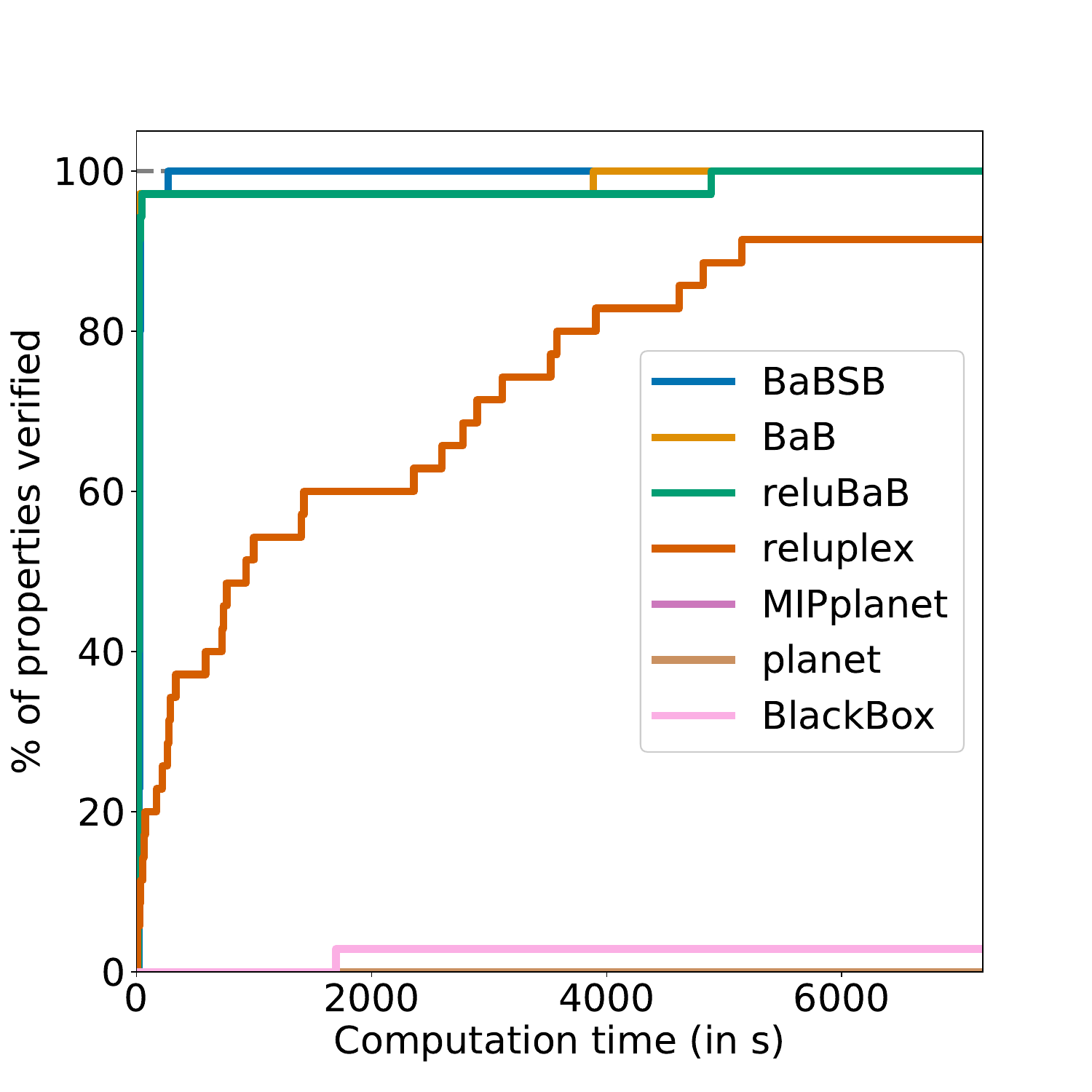}
      \caption{On SAT properties}
    \end{subfigure}%
    \begin{subfigure}[t]{.50\textwidth}
      \vskip 0pt
      \includegraphics[width=.90\textwidth]{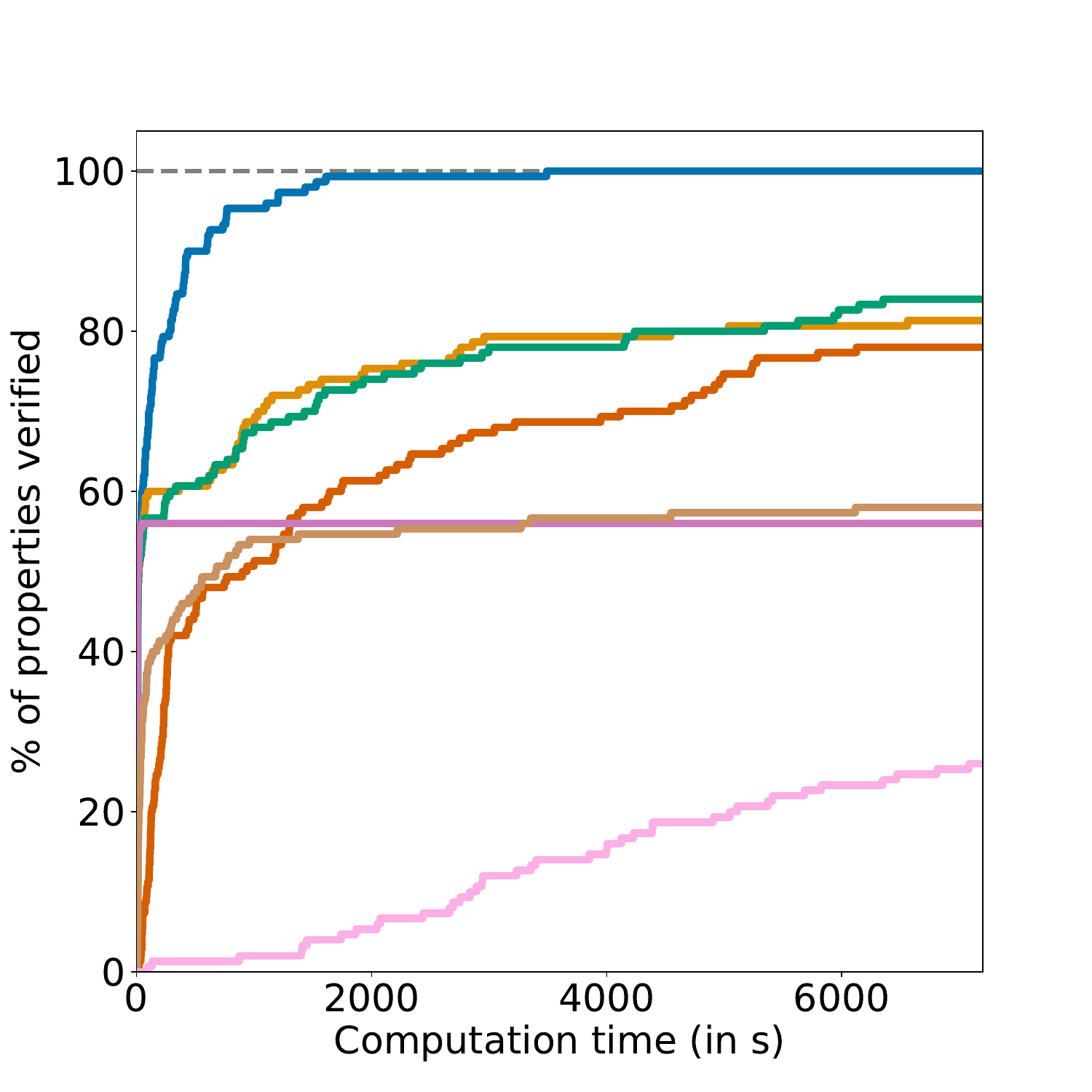}
      \caption{On UNSAT properties}
    \end{subfigure}
  \end{minipage}

  \begin{minipage}[t]{\textwidth}
    \vskip 0pt
    \caption{\label{fig:sat_unsat_acas}  Proportion of properties
        verifiable by different methods under varying time budgets on
        the \textbf{ACAS} data set. We observe that \textbf{planet} doesn't succeed in
        solving any of the SAT properties, while our proposed methods are
        extremely efficient at it, even if there remains some properties that
        they can't solve.}
  \end{minipage}
  \vskip -20pt
\end{figure}

\begin{figure}
  \begin{minipage}[b]{\textwidth}
    \vskip -10pt
    \centering
    \begin{subfigure}[t]{.50\textwidth}
      \vskip 0pt
      \includegraphics[width=.90\textwidth]{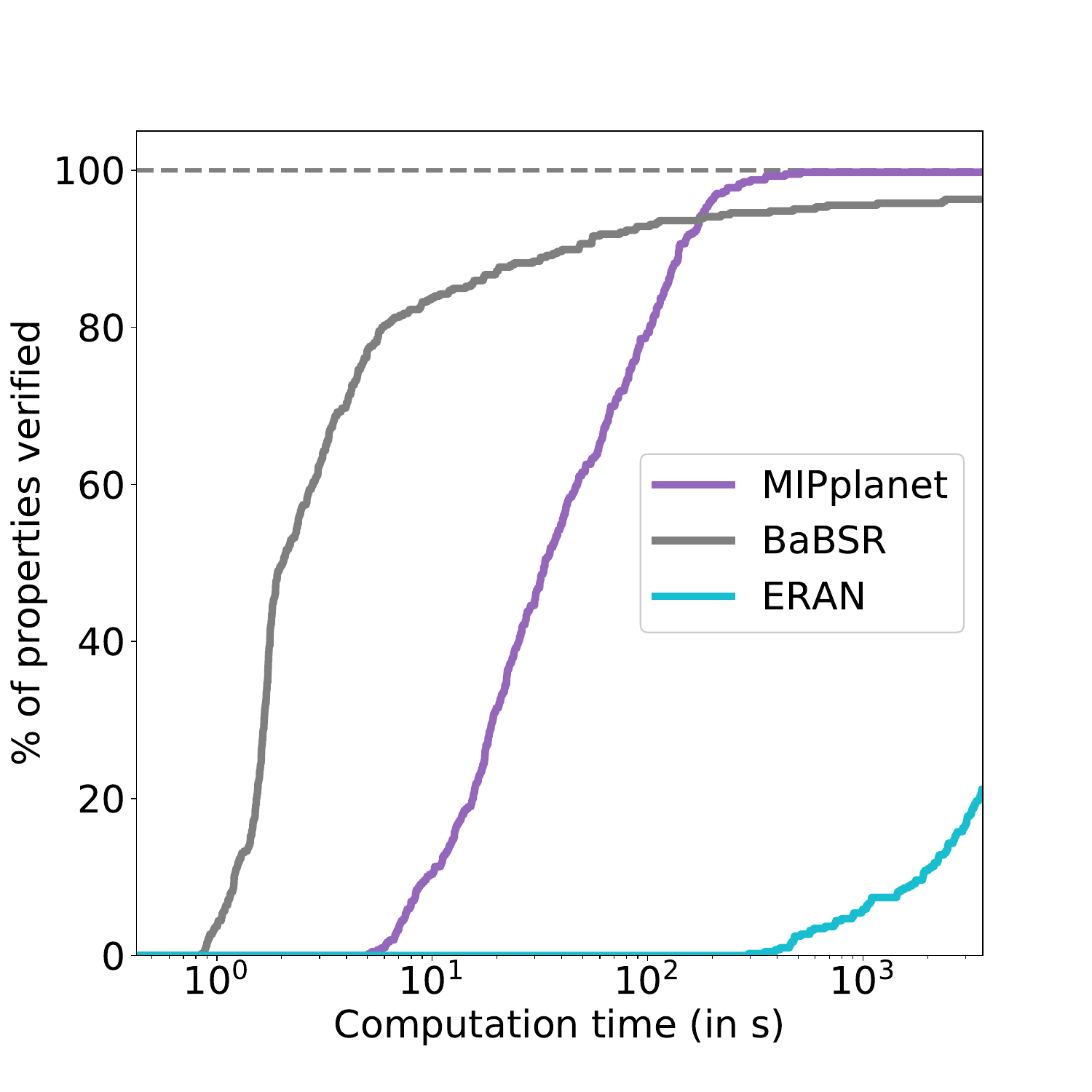}
      \caption{On SAT properties}
    \end{subfigure}%
    \begin{subfigure}[t]{.50\textwidth}
      \vskip 0pt
      \includegraphics[width=.90\textwidth]{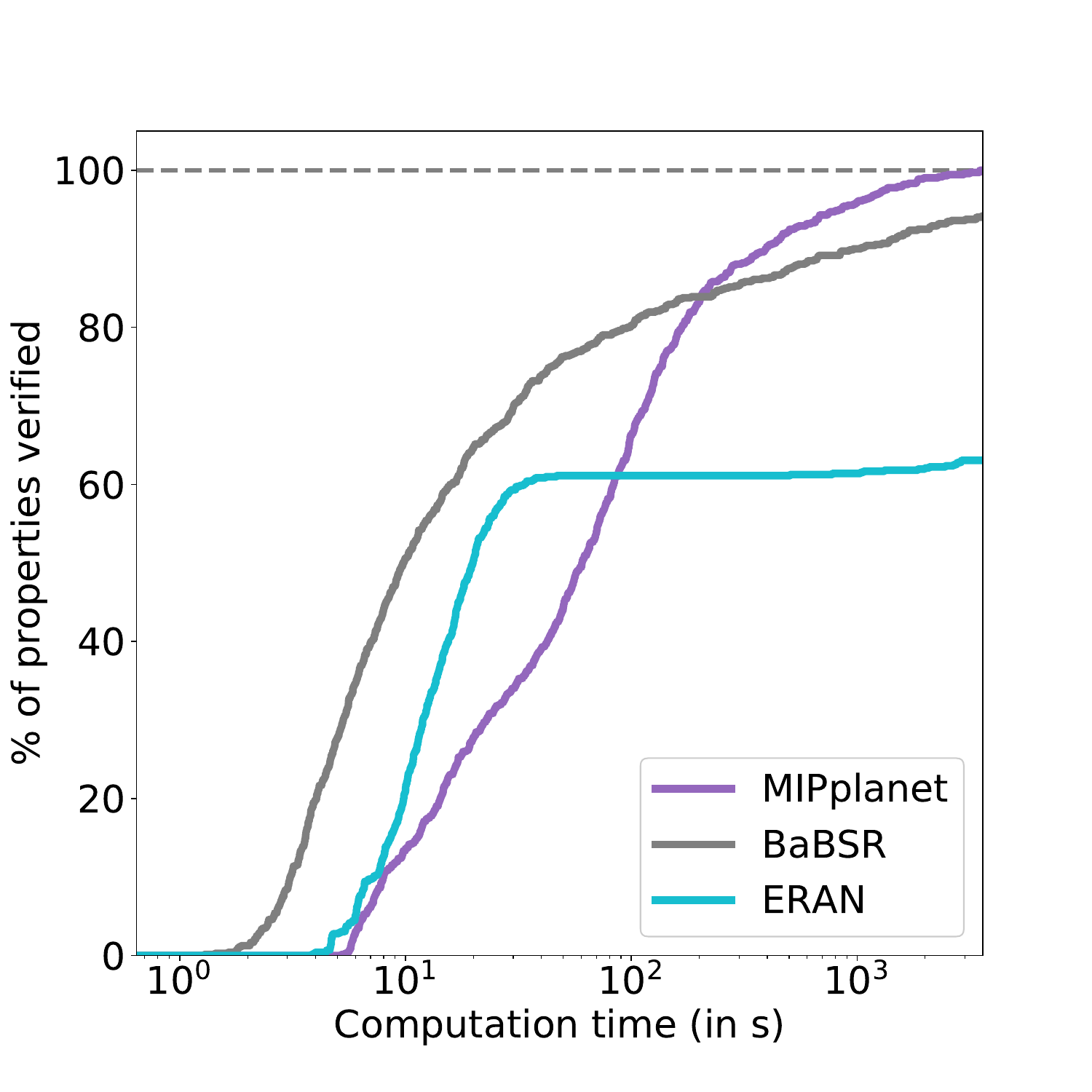}
      \caption{On UNSAT properties}
    \end{subfigure}
  \end{minipage}

  \begin{minipage}[t]{\textwidth}
    \caption{\label{fig:sat_unsat_med}Proportion of properties
        verifiable by different methods under varying time budgets on
        the \textbf{reduced Robust network} data set. Similar performances can be observed on both SAT and UNSAT properties. In terms of the total number of properties solved, \textbf{MIPplanet} slightly outperforms \textbf{BaBSR} on challenging UNSAT problems. However, on simple problems, \textbf{BaBSR} are much more time efficient than \textbf{MIPplanet}.}
  \end{minipage}
    \vskip -20pt
\end{figure}

\begin{figure}
  \begin{minipage}[b]{\textwidth}
    \vskip -10pt
    \centering
    \begin{subfigure}[t]{.50\textwidth}
      \vskip 0pt
      \includegraphics[width=.90\textwidth]{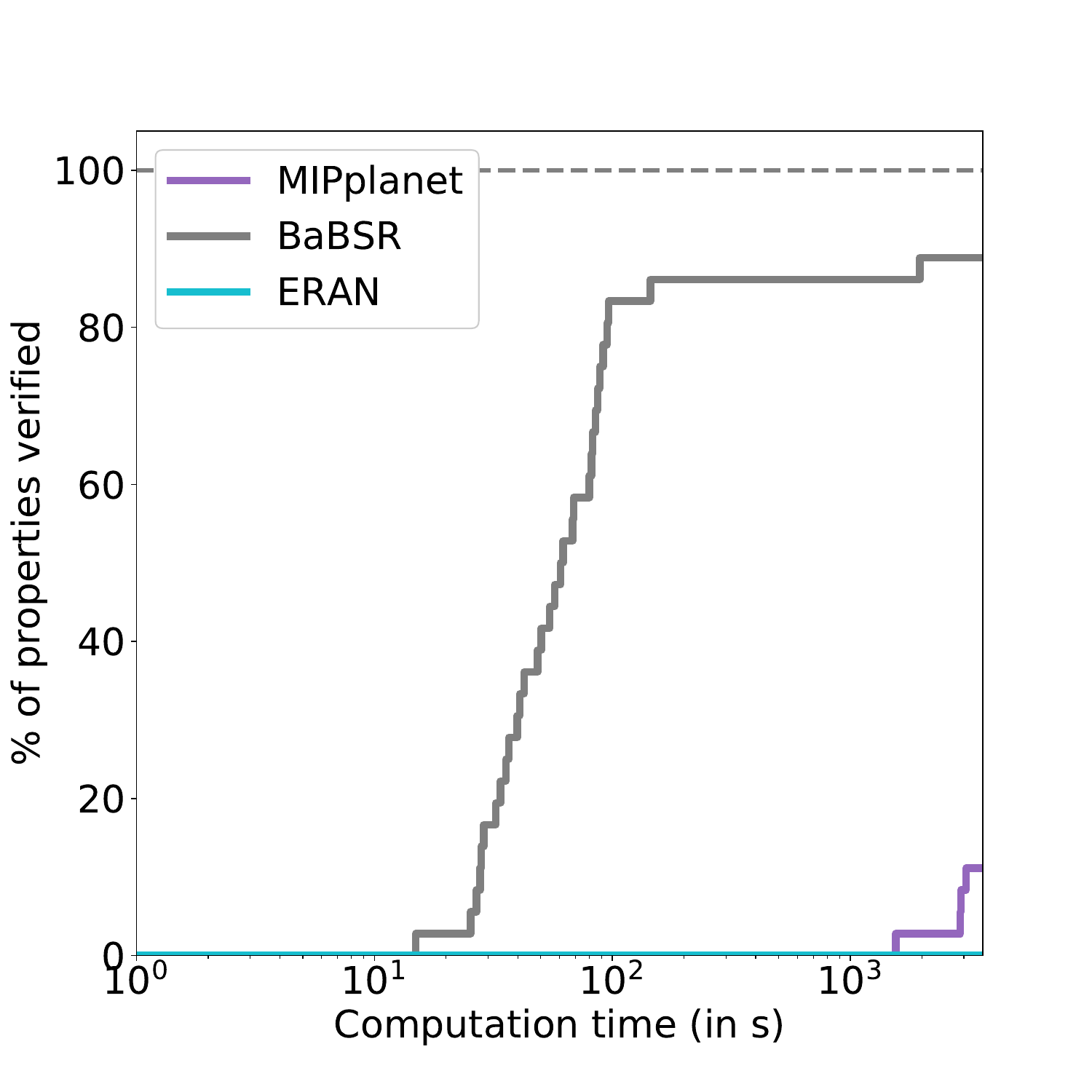}
      \caption{On SAT properties}
    \end{subfigure}%
    \begin{subfigure}[t]{.50\textwidth}
      \vskip 0pt
      \includegraphics[width=.90\textwidth]{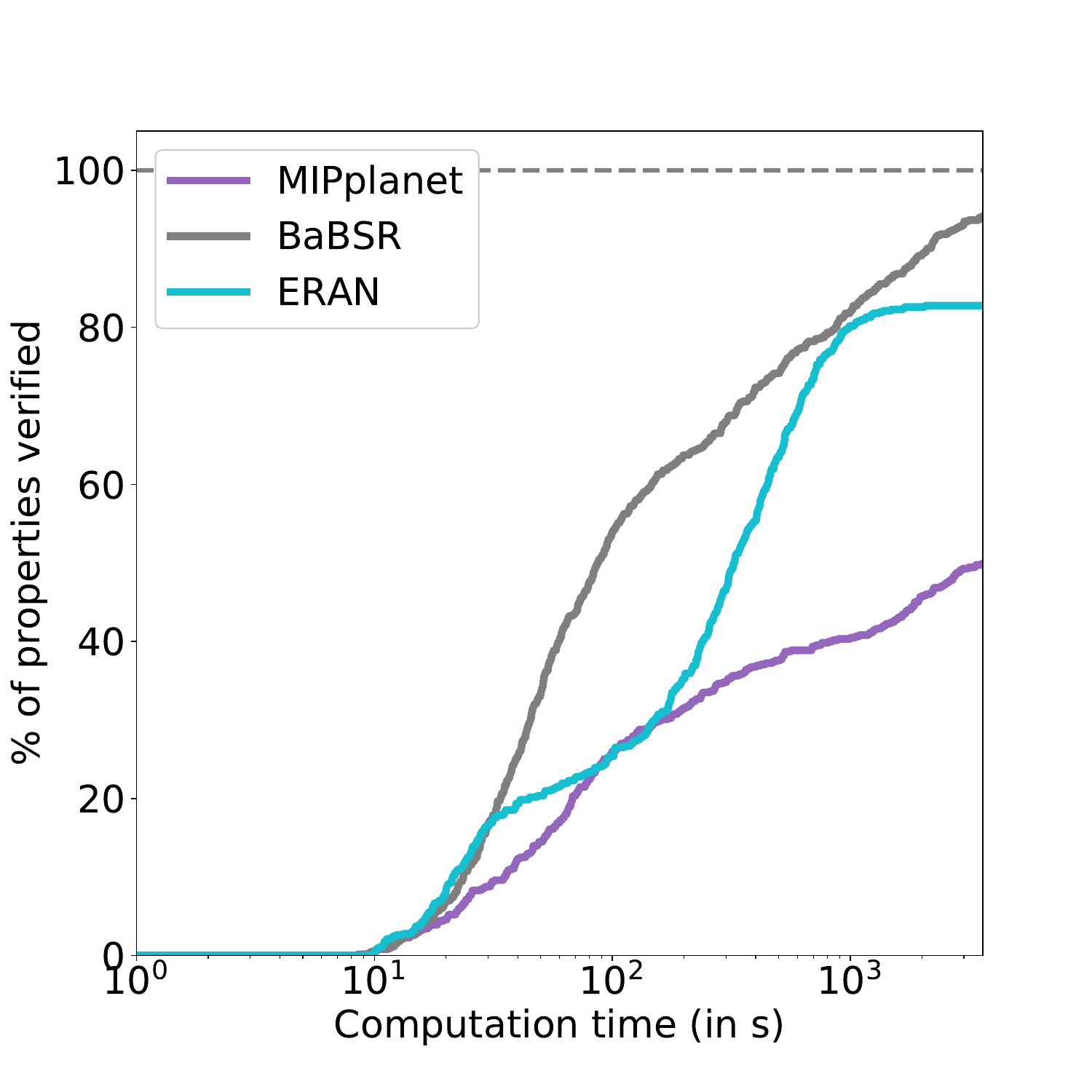}
      \caption{On UNSAT properties}
    \end{subfigure}
  \end{minipage}

  \begin{minipage}[t]{\textwidth}
    \vskip 0pt
    \caption{\label{fig:sat_unsat_large} Proportion of properties
        verifiable by different methods under varying time budgets on
        the \textbf{Robust Network} data set. \textbf{BaBSR} outperforms \textbf{MIPplanet} significantly in both cases. The huge performance gap on SAT properties indicates that Branch-and-Bound is an effective algorithm for finding counterexamples on large networks.}
  \end{minipage}
\end{figure}
\clearpage
\bibliography{shortstrings,oval,paper_specific}


\end{document}